%% file: arxiv.tex
\documentclass{article}

 \usepackage[preprint]{neurips_2026}

\usepackage[utf8]{inputenc} %
\usepackage[T1]{fontenc}    %
\usepackage{hyperref}       %
\usepackage{url}            %
\usepackage{booktabs}       %
\usepackage{amsfonts}       %
\usepackage{nicefrac}       %
\usepackage{microtype}      %
\usepackage{xcolor}         %

\usepackage{hyperref}
\usepackage{url}

\usepackage{booktabs}
\usepackage{multirow}
\usepackage{adjustbox}
\usepackage{amsmath}
\usepackage{makecell}

\usepackage{titletoc}

\usepackage{xstring}
\newcommand{\accdelta}[2]{#1\textsuperscript{\scriptsize\IfBeginWith{#2}{-}{\textcolor{red!70!black}{#2}}{\textcolor{green!50!black}{#2}}}}

\usepackage{graphicx}

\definecolor{kh}{HTML}{168aff}

\newcommand{\repourl}{\url{https://github.com/Chenghao-Qiu/Task-Preserving-ICL}}

\title{\input{sec/title}}

\author{
\begin{tabular}{ccc}
Chenghao Qiu & Chunli Peng & Yufeng Yang \\
\texttt{chenghaoqiu@tamu.edu} &
\texttt{chunli.peng@tamu.edu} &
\texttt{ynyang94@tamu.edu} \\[0.8em]
\multicolumn{3}{c}{
\begin{tabular}{cc}
Kuan-Hao Huang & Yi Zhou \\
\texttt{khhuang@tamu.edu} &
\texttt{yi.zhou@tamu.edu}
\end{tabular}
}
\end{tabular}
\\[3em]
Texas A\&M University
}

\begin{document}

\maketitle

\input{sec/abstract_arxiv}
\input{sec/1_intro}
\input{sec/2_related}
\input{sec/3_method}

\input{sec/4_experiment}

\input{sec/5_conclusion}

\clearpage
\bibliography{sec/reference}
\bibliographystyle{plainnat}

\clearpage
\appendix
\input{sec/appendix_arxiv}

\end{document}

%% file: sec/title.tex
When Correct Demonstrations Hurt: Rethinking the Role of Exemplars in In-Context Learning

%% file: sec/abstract_arxiv.tex
\begin{abstract}
In-context learning (ICL) is often motivated by the intuition that demonstrations help because they provide correct input--output examples. However, we reveal a counterintuitive phenomenon: correctness does not guarantee exemplar utility, and some correct demonstrations can even reduce ICL accuracy.
To study this correctness--utility gap, we introduce task preserving perturbations, where only the exemplar input is changed, while the example remains a correct instance of the same task. Concretely, each perturbed exemplar is assigned the target induced by the task mapping.
This framework covers both label updating perturbations, where task relevant semantics change and targets are recomputed, and stricter target preserving perturbations, where the original target remains valid.  
We formalize the resulting failure mode as contextual evidence shift: task preserving perturbations can change the effective mixture of evidence used by the model for contextual inference, thereby separating exemplar correctness from exemplar utility.
Across sentiment classification, logical reasoning, and math word problems, we find that task preserving perturbed demonstrations can substantially degrade ICL performance, especially for smaller models, harder tasks, and higher perturbation ratios. 
Our results show that robust ICL requires evaluating not only whether demonstrations are correct, but also how they influence contextual inference.
Code is available at \repourl.
\end{abstract}

%% file: sec/1_intro.tex
\section{Introduction}

\begin{figure}[ht]
    \centering
    \includegraphics[width=\linewidth]{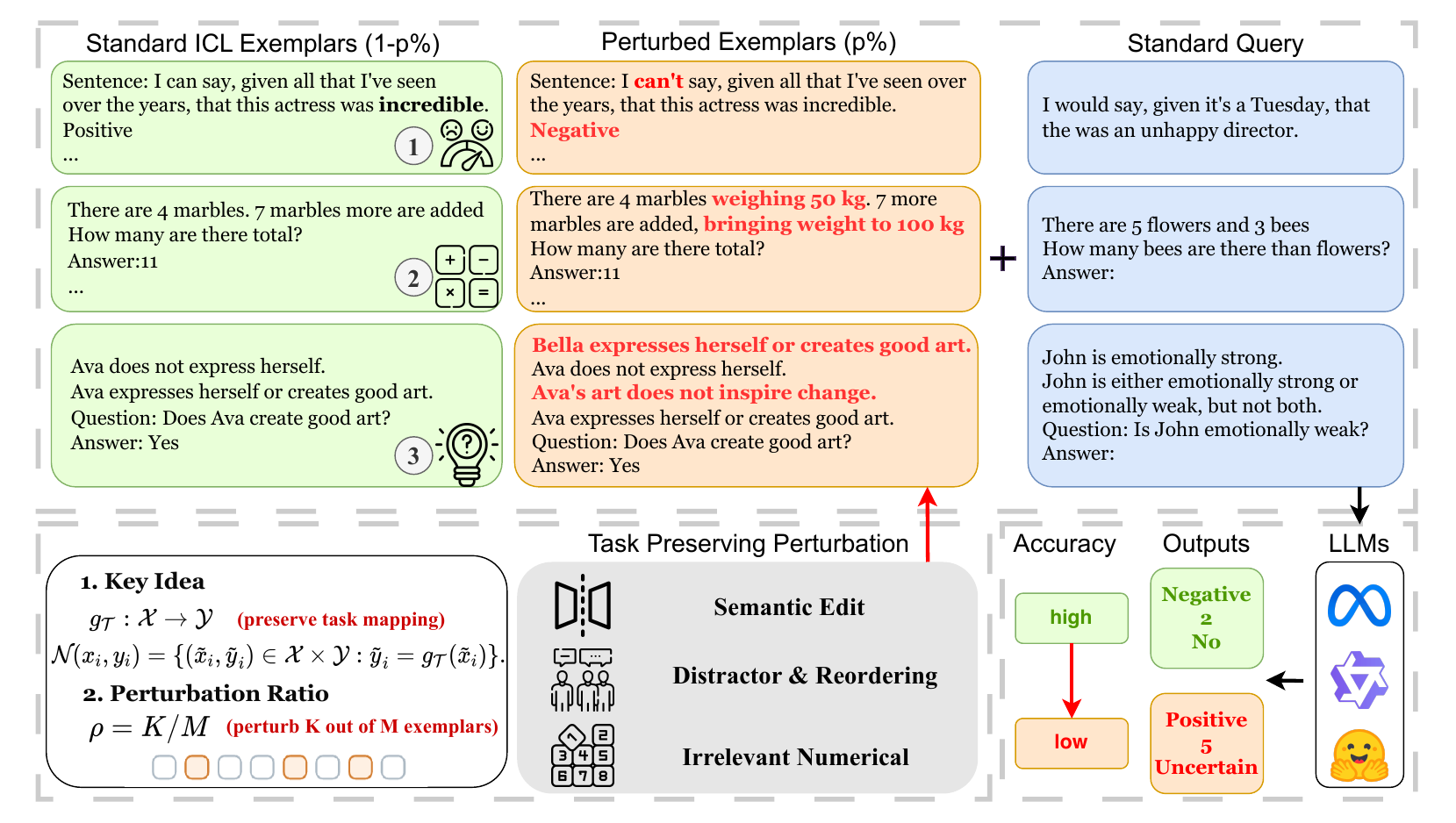}
    \caption{\textbf{Overview of Task Preserving Exemplar Perturbations.} We study task preserving exemplar perturbations across (1) sentiment analysis, (2) logical reasoning, and (3) math word tasks. Top: Exemplar construction under perturbation ratio $\rho$, where a proportion $\rho$ of exemplars is randomly selected for perturbation while the remaining exemplars are kept unchanged. Green denotes \textcolor{green!70!black}{original exemplars}, orange denotes \textcolor{orange!90!black}{perturbed exemplars}, and red highlights the \textcolor{red!90!black}{modified input tokens}.
    Bottom: Illustration of task preserving perturbations and their effect on in-context learning. Gray panels present \textcolor{gray}{concrete task preserving instantiations} for different tasks. Despite preserving the task mapping, perturbed exemplars can shift contextual evidence and induce erroneous model predictions.}
    \label{fig:overview}
\end{figure}

In-context learning (ICL) enables large language models (LLMs) to adapt to new tasks by conditioning on a small set of input--output demonstrations in the prompt, without updating model parameters~\cite{brown2020language}. 
A common intuition is that demonstrations help because they provide correct examples: each exemplar input is paired with the output induced by the target task mapping.
However, prior analyses suggest that demonstrations do not simply specify a mapping~\cite{min2022rethinking}, they also communicate the label space~\cite{wei2023larger}, input distribution~\cite{zhou2023hijacking}, and prompt format. Their effectiveness further depends on exemplar ordering~\cite{lu2022fantastically}, selection~\cite{liu2022makes}, and retrieval quality~\cite{rubin2022learning}.
These findings show that demonstrations affect ICL through factors beyond the input--output mapping, but they do not directly test whether task correct demonstrations can become harmful.
This raises a more basic question: \emph{when a demonstration is correct, under what conditions does it actually help?}

In this work, we argue that \emph{exemplar correctness is not sufficient for exemplar utility}. 
A demonstration can be individually valid under the task definition while still being harmful as contextual evidence. 
For instance, a sentiment exemplar such as ``I can't say this film is wonderful'' $\rightarrow$ negative is correct under the sentiment classification task. However, when used as an in-context exemplar for a query such as ``I can say this is a beautiful place,'' it may provide little useful evidence or even introduce misleading contextual evidence that shifts the model away from the intended prediction.
The harm arises not from label incorrectness, but from a mismatch between the exemplar's input regime and the intended query distribution, which can shift the model's inferred decision context.
This distinction matters because ICL pipelines typically start from correct candidate exemplars, but exemplar correctness is a local property of an input--output pair. Exemplar utility, by contrast, is context dependent: it depends on how an exemplar interacts with other exemplars and with the intended query distribution. 
Rather than separately utilizing each exemplar, the model uses the full exemplar set to infer the latent task, input regime, and decision rule.
Under this view, correct exemplars can still support competing contextual hypotheses: they may preserve the same task mapping while shifting surface form or input distribution enough to change model behavior.
Existing studies do not fully isolate this correctness--utility gap. Label space perturbation studies test whether models can learn a new mapping from context~\cite{wei2023larger,shi2024larger}, but intentionally alter the supervision signal carried by demonstrations. Input space adversarial studies show that demonstrations form an effective attack surface~\cite{wang2023adversarial,zhou2023hijacking}, but they often steer the model toward attacker chosen outputs or even alter task relevant semantics while retaining the original label~\cite{chu2026bamicl}. As a result, prior work often conflates whether demonstrations remain correct for the task with whether correct demonstrations remain useful once their inputs are expressed differently.

To isolate the distinction between exemplar correctness and exemplar utility, we propose \emph{task preserving perturbations} as a controlled input side intervention on in-context exemplars. Figure~\ref{fig:overview} provides a schematic illustration of the exemplar construction process under perturbation ratio $\rho$ and how task preserving perturbations affect downstream predictions. By task preserving, we only modify selected exemplar inputs while maintaining output space, task mapping, and exemplar order. Each transformed exemplar is then paired with the target induced by the same task mapping. This framework covers two regimes: label updating perturbations, where task relevant semantics change and the target is recomputed, and stricter target-preserving perturbations, where the original label or answer remains valid. In this way, the perturbations remain task correct while altering the contextual evidence available to the model. 
Across sentiment classification, logical reasoning, and math word problems, we find that task preserving perturbed exemplars can substantially reduce ICL performance, particularly for smaller models, harder tasks, and larger perturbation ratios. On SST-2, replacing clean exemplars with task preserving alternatives consistently degrades accuracy across model families and scales, in some cases making performance even worse than zero-shot prompting.
Further controls and matched distribution evaluations suggest that the degradation reflects a shift in contextual evidence, rather than generic task recognition failure or a failure to interpret the perturbed inputs.
Additional analyses based on original perturbed similarity, exemplar position, and attention allocation further support the view that correct exemplars can become harmful by changing the evidence mixture used for contextual inference.
We summarize our contributions as follows:
\begin{itemize}
    \item We identify a correctness-utility gap in ICL that demonstrations can remain valid under the same task mapping while still degrading downstream performance.

    \item Methodologically, we introduce task preserving perturbations, a controlled input side intervention that modifies only exemplar inputs while preserving the task mapping. This effect is formalized as contextual evidence shift, showing how such perturbations can alter the effective evidence mixture used for contextual inference.

    \item Extensive experiments across multiple tasks, model families, and model scales show that task preserving perturbed demonstrations can substantially degrade ICL performance. Additional analyses further support contextual evidence shift as an explanation for this failure.
\end{itemize}

%% file: sec/2_related.tex
\section{Related Work}
\subsection{In-Context Learning}
As large language models (LLMs) have become general-purpose foundation models~\cite{achiam2023gpt,guo2025deepseek,chen2021evaluating,roziere2023code}, in-context learning (ICL) has emerged as a central capability: a model can perform a task by conditioning on a few demonstrations in the prompt, without any parameter updates~\cite{brown2020language}.
Early studies~\cite{min2022rethinking} on ICL mainly focused on what information is conveyed by demonstrations. Following this path, \cite{wei2023larger} use label perturbations to show that larger models can infer new task mappings from context. \cite{shi2024larger} further attribute this scale effect to larger models' ability to use broader feature sets. 
Meanwhile, input side perturbation studies~\cite{wang2023adversarial} show that adversarially perturbed demonstrations can substantially change model predictions. \cite{zhou2023hijacking} and \cite{chu2026bamicl} further study how malicious or budgeted demonstration manipulations can hijack ICL behavior. These works reveal that ICL is highly sensitive to demonstration level interventions. 
One line of theoretical work views ICL as implicit optimization inside the Transformer forward pass, where demonstrations serve as training examples that allow models to learn function classes in context~\cite{garg2022can} or approximate learning algorithms, including gradient-descent-like updates~\cite{von2023transformers,ahn2023transformers}, ridge regression~\cite{akyurek2023learning}, and more general forms of in-context algorithm selection~\cite{bai2023transformers}.
The second line interprets ICL through a kernel or Bayesian inference perspective. \cite{han2025understanding} show that ICL exhibits kernel regression like behavior, while \cite{panwar2024incontext} show that high-capacity Transformers can behave like Bayesian predictors over latent tasks. \cite{raventos2023pretraining} further show that pretraining task diversity modulates this behavior, shifting models from Bayesian like estimators under limited diversity toward ridge regression style predictors under broader diversity.

\subsection{Demonstration Selection}

A growing line of work studies why some demonstrations are more useful than others in ICL. 
~\cite{lu2022fantastically} show that the order of in-context examples can substantially affect model predictions. 
\cite{liu2022makes} utilize similarity based selection to identify examples that better match the test query, while retrieval based prompting~\cite{rubin2022learning} learns to retrieve useful demonstrations for downstream prediction. 
Beyond similarity, selective annotation and active selection methods show that representative, diverse, or model informative examples can yield stronger few-shot performance than randomly chosen demonstrations~\cite{su2023selective,zhang2022active}. 
Coverage oriented selection methods further suggest that the utility of a demonstration set depends on its global composition rather than only on individual example quality~\cite{gupta2023coverage,li2023finding}. 
Recent influence based analyses also suggest that in-context examples can have heterogeneous effects on model predictions~\cite{nguyen2023context}. 

%% file: sec/3_method.tex
\section{Method}
\label{sec:method}

\subsection{ICL Setup}

Let $\mathcal{T}$ be a task with input space $\mathcal{X}$ and output space $\mathcal{Y}$. An ICL prompt contains $M$ demonstrations
\begin{equation}
D = \bigl((x_1,y_1),\ldots,(x_M,y_M)\bigr), \qquad (x_i,y_i) \in \mathcal{X} \times \mathcal{Y},
\end{equation}
followed by a query input $x^{\mathrm{q}}$. Under a fixed prompt template $\Pi(\cdot)$, a language model $f_\theta$ predicts
\begin{equation}
\hat{y} = f_\theta\bigl(\Pi(D, x^{\mathrm{q}})\bigr).
\end{equation}
For classification tasks, $y \in \mathcal{Y}$ is a class label; for reasoning tasks, $y$ denotes a normalized answer string. Throughout, we keep the instruction text, label names, demonstration order, and query input fixed unless otherwise stated. The only attack surface is the \emph{input side} of the demonstrations.

\subsection{Task Preserving Perturbations}
We study task preserving perturbations on the input side of in-context exemplars. Let $g_{\mathcal{T}}:\mathcal{X}\rightarrow\mathcal{Y}$ denote the gold input--output mapping induced by task $\mathcal{T}$. Given an original exemplar $(x_i,y_i)$ with $y_i=g_{\mathcal{T}}(x_i)$, a perturbation maps the exemplar input $x_i$ to $\tilde{x}_i$. The corresponding target is then determined by the same task mapping as $\tilde{y}_i=g_{\mathcal{T}}(\tilde{x}_i)$. We define the admissible set of task preserving perturbations as
\begin{equation}
\mathcal{N}_{\mathrm{task}}(x_i,y_i)
=
\left\{
(\tilde{x}_i,\tilde{y}_i)\in \mathcal{X}\times\mathcal{Y}
:
\tilde{y}_i = g_{\mathcal{T}}(\tilde{x}_i)
\right\}.
\end{equation}
This condition keeps the task definition, prompt format, demonstration order, and query input fixed, while requiring every perturbed exemplar to remain a valid input output pair under the same task. Importantly, task consistency does not require the output token to remain identical. Target updates are required only for perturbations that change task relevant semantics, whereas semantics preserving perturbations retain the original target.
Within this umbrella, strict label preserving perturbation is a special case where the transformed input must preserve the original exemplar target
\begin{equation}
\mathcal{N}_{\mathrm{label}}(x_i,y_i)
=
\left\{
(\tilde{x}_i,y_i)\in \mathcal{X}\times\mathcal{Y}
:
g_{\mathcal{T}}(\tilde{x}_i)=y_i
\right\}.
\end{equation}
Thus, task preserving perturbations allow the target to be recomputed under the fixed task mapping, whereas strict label preserving perturbations additionally require maintaining the original target. This distinction isolates input side perturbations from label corruption, because each perturbed exemplar is paired with the correct output under $g_{\mathcal{T}}$ rather than an adversarially chosen target.

In our experiments, we instantiate this framework in different ways across tasks. For sentiment analysis, we adopt the label updating regime. A sentence is edited so that its sentiment changes, and the label is updated according to the same sentiment classification task. For example, \textit{``I can say, given my experience, that this film is wonderful.''} with label \textit{positive} can be changed into \textit{``I can't say, given my experience, that this film is wonderful.''} with label \textit{negative}. 
For logical reasoning tasks, we use a stricter target preserving regime. The original facts, CoT reasoning, and gold answer remain unchanged, while distracting facts are added and the premises are reordered. For example, an original statement such as \textit{``Alice is a famous musician''} is kept, while an additional statement such as \textit{``Bob is a famous musician''} is added and follows by premise shuffling. 
Similarly, for math word problems, irrelevant numerical information is added to the problem statement, while the question, reasoning steps, and final answer remain unchanged. For example, \textit{``There are 4 marbles. 7 marbles more are added.''} can be changed into \textit{``There are 4 marbles weighing 50 kg. 7 more marbles are added, bringing the total weight to 100 kg.''} under the same question \textit{``How many are there total?''}.

\subsection{Perturbation Budget}
\label{sec:budget_placement}
Given $M$ in-context exemplars and a perturbation budget $K\le M$, we define the perturbation ratio as $\rho=K/M$. A placement policy $\pi$ selects the perturbed index set
\begin{equation}
I_{\pi,K}=\pi(M,K)\subseteq \{1,\ldots,M\},
\qquad |I_{\pi,K}|=K.
\end{equation}
Unless otherwise specified, $I_{\pi,K}$ is sampled uniformly at random. The resulting perturbed context is

\begin{equation}
\tilde{D}_{\rho,\pi}
=
\left\{
(\tilde{x}_i,\tilde{y}_i)\ \text{if } i\in I_{\pi,K},
\ \text{otherwise } (x_i,y_i)
\right\}_{i=1}^{M}.
\end{equation}

\subsection{Contextual Evidence Shift}
\label{sec:contextual_evidence_shift}

To explain why task preserving perturbations can make correct exemplars harmful, we interpret them as inducing a shift in the contextual evidence used by ICL. 
Let $P_0$ and $P_1$ denote the clean and perturbed input regimes, respectively. Both regimes share the same task mapping $g_{\mathcal{T}}:\mathcal{X}\rightarrow\mathcal{Y}$, but induce different task valid exemplar distributions:
\begin{equation}
\mathcal{P}^{\mathcal{T}}_0(x,y)=P_0(x)\delta_{g_{\mathcal{T}}(x)}(y),
\qquad
\mathcal{P}^{\mathcal{T}}_1(x,y)=P_1(x)\delta_{g_{\mathcal{T}}(x)}(y).
\label{eq:clean-perturbed-regimes}
\end{equation}
Thus, clean and perturbed exemplars differ in input regimes, while remaining valid under the same task mapping. At the distributional level, the contextual evidence induced by a prompt with perturbation ratio $\rho$ can be idealized as
\begin{equation}
\mathcal{P}^{\mathcal{T}}_{\rho}
=
(1-\rho)\mathcal{P}^{\mathcal{T}}_0
+
\rho\mathcal{P}^{\mathcal{T}}_1.
\end{equation}
Such a mixture changes the empirical evidence from which the model infers how the query should be interpreted.
This view is consistent with Bayesian and kernel-based interpretations of ICL, where the model behaves as a predictor induced by contextual examples. We write the prediction as
\begin{equation}
p_{\theta}(y\mid x_q,D_{\rho})
\approx
\int p_{\theta}(y\mid x_q,h)q_{\theta}(h\mid D_{\rho})dh,
\label{eq:posterior-predictive}
\end{equation}
where $h$ denotes the latent contextual hypothesis inferred from the prompt. In our setting, $h$ may encode not only the task mapping, but also assumptions about the input regime, salient features, distractor handling, and solution format. Task preserving perturbations keep $g_{\mathcal{T}}$ valid while shifting $q_{\theta}(h\mid D_{\rho})$ toward the perturbed regime. 

Let $\ell(\hat{y},y)$ denote an evaluation loss, instantiated as zero-one error for classification and exact-match error for reasoning tasks. For a fixed demonstration set $D$, we define the expected risk on an evaluation distribution $Q$ as
\begin{equation}
\mathcal{R}_{Q}(f_{\theta}(\cdot\mid D))
=
\mathbb{E}_{(x,y)\sim Q}
\left[
\ell\!\left(f_{\theta}(\Pi(D,x)),y\right)
\right].
\label{eq:expected-risk}
\end{equation}

We now formalize the distinction between exemplar correctness and exemplar utility. Correctness is an exemplar level condition requiring the target of each exemplar pair to be induced by the task mapping. Utility, by contrast, is context and distribution dependent. We write:
\begin{equation}
\mathrm{Correctness:}\quad
y_i = g_{\mathcal{T}}(x_i).
\label{eq:exemplar-correctness}
\end{equation}
\begin{equation}
\mathrm{Utility:}\quad
\Delta \mathcal{R}_{Q}(D,i)
=
\mathcal{R}_{Q}\!\left(f_{\theta}(\cdot\mid D\setminus i)\right)
-
\mathcal{R}_{Q}\!\left(f_{\theta}(\cdot\mid D)\right).
\label{eq:exemplar-utility}
\end{equation}
Under this convention, $\Delta \mathcal{R}_{Q}(D,i)>0$ indicates that the exemplar reduces risk, while $\Delta \mathcal{R}_{Q}(D,i)<0$ indicates that it is harmful. This formulation captures the central correctness-utility gap that a demonstration can satisfy Eq.~\eqref{eq:exemplar-correctness} while having negative utility under Eq.~\eqref{eq:exemplar-utility}. 
In particular, a perturbed exemplar drawn from $\mathcal{P}^{\mathcal{T}}_1$ can be correct under $g_{\mathcal{T}}$ but have negative utility for clean queries evaluated under $Q=\mathcal{P}^{\mathcal{T}}_0$.
For example, in sentiment analysis, both $\mathcal{P}^{\mathcal{T}}_0$ and $\mathcal{P}^{\mathcal{T}}_1$ are correct, but they may provide different contextual cues. 
Negation based edits in $\mathcal{P}^{\mathcal{T}}_1$ can make polarity reversal salient in the prompt. 
When many exemplars exhibit this pattern, the model may treat it as part of the contextual decision rule and apply it to clean queries from $\mathcal{P}^{\mathcal{T}}_0$ with similar surface forms.
This explains why clean test performance can degrade as $\rho$ increases. 
We provide a detailed derivation in Appendix~\ref{app:effective_evidence_mass} about effective perturbed evidence mass $m_{\theta}(D_{\rho},x_q)$ and its connection to utility.

%% file: sec/4_experiment.tex
\section{Experimental Results}
\label{sec:experiment}

\paragraph{Models.}
We evaluate open weight large language models from several recent generations, including Llama-2~\cite{touvron2023llama}, Llama-3.1~\cite{grattafiori2024llama}, Qwen2.5~\cite{yang2024qwen2}, Qwen3.5~\cite{qwen3.5}, and Gemma-4. These families cover models released across different stages of recent LLM development, enabling us to study whether the effect of task preserving exemplar perturbations is consistent across model families, scales, and model generations. Unless otherwise specified, all evaluated models are instruction-tuned or chat-tuned variants. Detailed model identifiers and access information are provided in Appendix~\ref{app:model_details}.

\paragraph{Datasets.} 
We utilize human crafted examples from AdvGLUE~\cite{wang2adversarial}, where the semantic content of both the original and perturbed inputs is interpretable to human readers. For logical reasoning tasks, we use ProverQA~\cite{qi2025large} to generate logical question answering instances with three levels of difficulty. For math reasoning tasks, we use PROBLEMATHIC~\cite{anantheswaran2025cutting}, a benchmark of simple and complex math word problems with both clean and perturbed variants. All exemplars are organized according to the formatting described in Appendix~\ref{app:exemplar_format}.

\paragraph{Perturbation Instantiations.}
All experiments follow the task preserving perturbation framework in Section~\ref{sec:method}. SST-2 instantiates the label updating regime, whereas ProverQA and PROBLEMATHIC adopt stricter label or answer preserving regimes. Instantiation details are provided in Appendix~\ref{app:perturbation_details}.

\paragraph{Evaluation Metrics.} 
To evaluate the performance degradation introduced by perturbed in-context exemplars, we report the mean accuracy averaged over all evaluation instances. For math tasks, we use \emph{Exact Match} (EM) as the evaluation metric, where a prediction is counted as correct only when the generated answer exactly matches the ground truth answer. 

\paragraph{Baselines.} Across all experiments in this section, we report both the zero-shot result and the 0\% perturbation condition. The latter corresponds to the clean ICL setting with fully unperturbed exemplars. These two baselines allow us to measure both the degradation relative to clean ICL and whether perturbed exemplars still remain helpful over zero-shot prompting.

\subsection{Sentiment Analysis}
To study the effect of task preserving perturbation on a relatively simple classification task, we perform experiments on SST-2 using perturbation ratios from 25\% to 100\% with the number of exemplars $M=32$ fixed. Because the perturbed exemplars change semantically, we update their labels accordingly to maintain the task mapping. 
To ensure that task preserving perturbations do not merely confuse the model about the task being performed, each exemplar and final query is accompanied by an explicit instruction \textit{``Instruction: You are doing sentiment analysis. Only output positive or negative.''}. We further introduce a task irrelevant control setting, where selected exemplar inputs are replaced with sentiment-neutral factual statements, such as \textit{``The sun rises in the east.''}. This control helps distinguish the effect of task preserving adversarial evidence from the generic effect of disrupting the model's task recognition. For every case, the replaced exemplars are randomly selected, and we run the experiment with 100 different random seeds. 

\begin{figure}[ht]
    \centering
    \includegraphics[width=\linewidth]{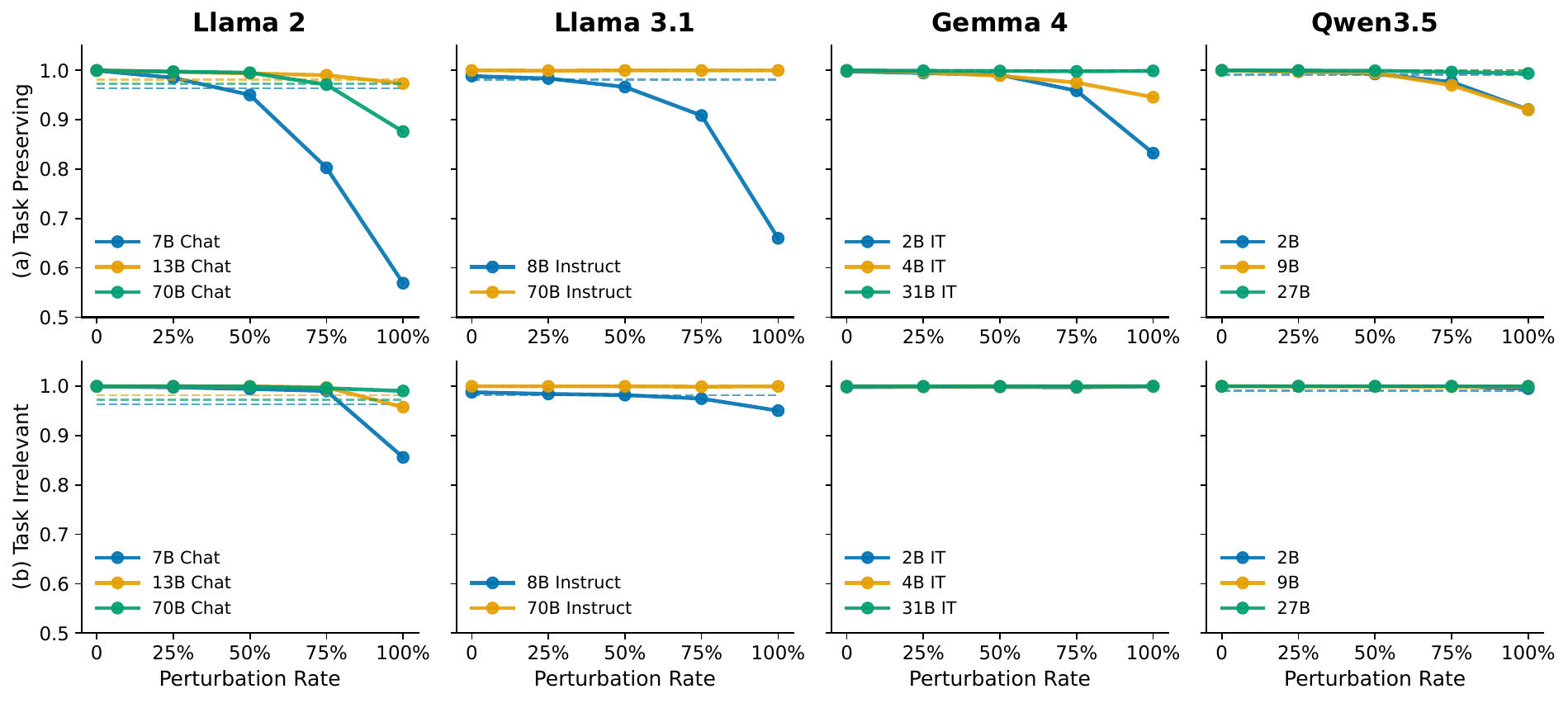}
    \caption{\textbf{Sentiment Analysis Performance.} 
    We evaluate SST-2 with 32 in-context exemplars under different exemplar perturbation ratios. 
    Top row: selected exemplars are replaced by task preserving perturbed exemplars constructed using our input side perturbation method. 
    Bottom row: selected exemplars are instead replaced by task irrelevant factual sentences.
    The x-axis shows the perturbation ratio, and the y-axis reports mean accuracy over 100 random runs. 
    Dashed lines denote the corresponding zero-shot accuracy on each evaluation set. 
    }
    \label{fig:sst2}
\end{figure}

\paragraph{Results.}
Figure~\ref{fig:sst2} shows two complementary effects of exemplar perturbations. 
Under task preserving perturbations, performance generally decreases as the perturbation ratio increases, but the magnitude of degradation depends strongly on model scale and model generation. Within the same model family, smaller models are generally more vulnerable than larger models. For example, \textsc{Llama-2-7B} drops from 96.4\% accuracy at 0\% perturbation to 56.9\% at 100\% perturbation, while \textsc{Llama-3.1-8B} decreases from 98.2\% to 66.0\%. By contrast, larger models in the same families remain substantially more stable. 
This scale dependent pattern is even more pronounced for newer model families. In \textsc{Gemma-4} and \textsc{Qwen-3.5}, the largest models are almost unaffected by task preserving perturbations, whereas the smaller variants still exhibit nontrivial degradation at high perturbation ratios. 
Meanwhile, comparing the top and bottom rows of Figure~\ref{fig:sst2} shows that the degradation cannot be explained simply by generic disruption of the prompt or by confusion about the task identity. In the task irrelevant control setting where selected exemplar inputs are replaced by sentiment-neutral factual statements, models show only mild degradation. 
The effect is especially limited for \textsc{Gemma-4} and \textsc{Qwen-3.5} where task irrelevant replacements have little or no visible impact. This contrast indicates that task preserving perturbations are not merely weakening the model's recognition that the task is sentiment analysis. Rather, they introduce task valid but distributionally shifted examples that alter the contextual evidence used to infer the decision boundary within the same task. Additional results are provided in Appendix~\ref{app:all_sentiment}.
This further confirms the correctness-utility gap: demonstrations can remain individually correct while still becoming harmful.

\begin{table}[ht]
\centering
\small
\setlength{\tabcolsep}{2.5pt}
\renewcommand{\arraystretch}{0.98}
\caption{\textbf{Logical Reasoning Performance.} We perform experiments on the ProverQA dataset~\cite{qi2025large} with 8 in-context exemplars under different perturbation ratios. Levels correspond to different numbers of logical reasoning hops. Accuracy (\%) is averaged over 5 random runs. For the 25\%, 50\%, 75\%, and 100\% columns, superscripts indicate the signed relative change (\%) with respect to the 0\% condition in the same row. For each difficulty level and perturbation ratio, the largest performance drop is highlighted in \textbf{bold}, and the second largest drop is \underline{underlined}. 
}
\label{tab:proverqa}
\resizebox{0.95\columnwidth}{!}{%
\begin{tabular*}{\columnwidth}{@{\extracolsep{\fill}}llrrrrrr}
\toprule
\textbf{Level} & \textbf{Model} & \textbf{Zero} & \textbf{0\%} & \textbf{25\%} & \textbf{50\%} & \textbf{75\%} & \textbf{100\%} \\
\midrule

\multirow{8}{*}{Medium}
& Llama-3.1-8B   & 57.3 & 72.9 & \accdelta{74.2}{+1.8} & \textbf{\accdelta{71.0}{-2.6}} & \underline{\accdelta{70.6}{-3.2}} & \textbf{\accdelta{68.1}{-6.6}} \\
& Llama-3.1-70B  & 74.3 & 87.2 & \accdelta{88.8}{+1.8} & \accdelta{88.0}{+0.9} & \accdelta{87.0}{-0.2} & \accdelta{86.0}{-1.4} \\
& Qwen2.5-14B    & 83.0 & 88.3 & \underline{\accdelta{87.0}{-1.5}} & \accdelta{86.9}{-1.6} & \accdelta{86.7}{-1.8} & \accdelta{85.5}{-3.2} \\
& Qwen2.5-72B    & 82.0 & 86.8 & \accdelta{86.0}{-0.9} & \accdelta{86.0}{-0.9} & \accdelta{85.8}{-1.2} & \accdelta{85.4}{-1.6} \\
& Gemma-4-4B     & 65.5 & 85.2 & \accdelta{84.1}{-1.3} & \textbf{\accdelta{83.0}{-2.6}} & \textbf{\accdelta{82.0}{-3.8}} & \underline{\accdelta{80.9}{-5.0}} \\
& Gemma-4-31B    & 90.2 & 91.5 & \accdelta{91.0}{-0.5} & \accdelta{90.5}{-1.1} & \accdelta{91.0}{-0.5} & \accdelta{90.8}{-0.8} \\
& Qwen3.5-9B     & 83.8 & 86.9 & \textbf{\accdelta{85.2}{-2.0}} & \underline{\accdelta{85.4}{-1.7}} & \accdelta{85.2}{-2.0} & \accdelta{85.2}{-2.0} \\
& Qwen3.5-27B    & 89.0 & 92.1 & \accdelta{91.8}{-0.3} & \accdelta{91.9}{-0.2} & \accdelta{91.7}{-0.4} & \accdelta{91.7}{-0.4} \\

\midrule
\multirow{8}{*}{Hard}
& Llama-3.1-8B   & 53.0 & 57.8 & \accdelta{58.3}{+0.9} & \underline{\accdelta{55.5}{-4.0}} & \underline{\accdelta{54.1}{-6.4}} & \textbf{\accdelta{51.9}{-10.2}} \\
& Llama-3.1-70B  & 58.3 & 75.0 & \underline{\accdelta{73.7}{-1.7}} & \accdelta{72.5}{-3.3} & \accdelta{70.6}{-5.9} & \accdelta{69.7}{-7.1} \\
& Qwen2.5-14B    & 56.5 & 74.5 & \textbf{\accdelta{71.1}{-4.6}} & \textbf{\accdelta{70.1}{-5.9}} & \textbf{\accdelta{68.6}{-7.9}} & \underline{\accdelta{67.6}{-9.3}} \\
& Qwen2.5-72B    & 59.3 & 75.2 & \accdelta{74.3}{-1.2} & \accdelta{74.2}{-1.3} & \accdelta{72.0}{-4.3} & \accdelta{71.5}{-4.9} \\
& Gemma-4-4B     & 61.5 & 70.2 & \accdelta{69.7}{-0.7} & \accdelta{70.8}{+0.9} & \accdelta{69.1}{-1.6} & \accdelta{70.0}{-0.3} \\
& Gemma-4-31B    & 87.3 & 85.9 & \accdelta{85.6}{-0.3} & \accdelta{85.6}{-0.3} & \accdelta{85.4}{-0.6} & \accdelta{84.6}{-1.5} \\
& Qwen3.5-9B     & 73.3 & 71.5 & \accdelta{71.0}{-0.7} & \accdelta{72.3}{+1.1} & \accdelta{70.6}{-1.3} & \accdelta{69.7}{-2.5} \\
& Qwen3.5-27B    & 82.0 & 87.1 & \accdelta{87.1}{+0.0} & \accdelta{86.3}{-0.9} & \accdelta{87.6}{+0.6} & \accdelta{86.6}{-0.6} \\

\bottomrule
\end{tabular*}
}
\end{table}

\subsection{Logical Reasoning Task}
To examine whether the same phenomenon persists in a more challenging reasoning setting, we conduct experiments on ProverQA with difficulty levels that correspond to different numbers of reasoning hops. We vary the perturbation ratios from 25\% to 100\% with the number of exemplars $M=8$ fixed.
Following~\cite{qi2025large}, we adopt perturbations that combine distraction design with premise shuffling. 
These perturbations make the context more distracting and harder to interpret while leaving the gold chain-of-thought and the final label unchanged. In every case, the perturbed exemplars are randomly selected, and we report results averaged over 5 random runs.

\paragraph{Results.}
Table~\ref{tab:proverqa} shows that the effect of task preserving perturbations on logical reasoning is strongly scale dependent. On both the Medium and Hard splits, performance generally decreases as the perturbation ratio increases, with smaller models showing larger drops. For instance, \textsc{Llama-3.1-8B} drops by 6.6\% at 100\% perturbation on the Medium split, whereas \textsc{Llama-3.1-70B} drops by only 1.4\%. On the Hard split, \textsc{Qwen2.5-14B} drops by 9.3\%, while \textsc{Qwen2.5-72B} drops by 4.9\%. Meanwhile, stronger recent larger models remain within approximately 2\% of their clean ICL performance. This indicates that larger models remain more stable under the same perturbation budget, whereas smaller models become increasingly susceptible to irrelevant contextual information once the perturbation ratio is high. Results across all levels and models are reported in Appendix~\ref{app:all_qa}.

\subsection{Math Word Problems}
To examine task preserving perturbations in numerical reasoning, we conduct experiments on PROBLEMATHIC~\cite{anantheswaran2025cutting}. The perturbations add irrelevant or distracting numerical information to the problem statement while preserving the original solution and final answer. 
We evaluate the \textsc{Llama-2} family on the Simple and Complex splits with $M=16$ in-context exemplars and perturbation ratios from 25\% to 100\%. For each condition, the perturbed exemplars are randomly selected, and Exact Match accuracy is averaged over 10 random runs. This setup tests whether ICL remains effective when the input context contains additional distracting information.

\begin{figure}[ht]
    \centering
    \vspace{-2mm}
    \includegraphics[width=0.9\linewidth]{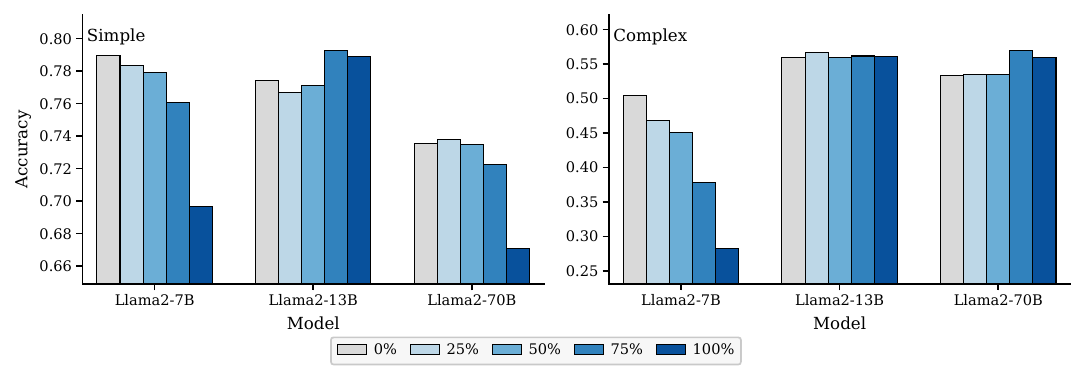}
    \vspace{-3mm}
    \caption{\textbf{Math Reasoning Performance.} We evaluate \textsc{Llama-2} models on the PROBLEMATHIC dataset~\cite{anantheswaran2025cutting} with 16 in-context exemplars under different input perturbation ratios denoted by different colors. The left and right panels report results on the Simple and Complex splits. Accuracy is measured by Exact Match (EM) and averaged over 10 runs. 
    }
    \label{fig:problemathic}
\end{figure}

\paragraph{Results.}
Figure~\ref{fig:problemathic} shows that task preserving perturbations on math word problems have a difficulty-dependent effect. On Simple split,  \textsc{Llama-2-7B} and \textsc{70B} both show noticeable degradation as the perturbation ratio increases, and model size does not yield a strictly monotonic advantage. 
Since these problems require only shallow arithmetic, this non-monotonic pattern should not be interpreted as evidence that smaller models reason better. Rather, irrelevant numerical distractors can make performance depend on format following and distractor filtering, not model scale alone~\cite{anantheswaran2025cutting,shi2024larger}.
On Complex split, only \textsc{7B} model shows a clear degradation across perturbation ratio, whereas larger models remain more stable. This suggests that, in more complex problems with perturbed contexts, smaller models struggle to separate task relevant quantities from irrelevant distractors, while larger models are better able to preserve the intended solution procedure.

\subsection{Perturbation Similarity Analysis}
\label{sec:similarity}

\begin{table*}[ht]
\centering
\small
\setlength{\tabcolsep}{4pt}
\renewcommand{\arraystretch}{1.08}
\caption{\textbf{Similarity Between Original and Perturbed Inputs.}
We summarize the similarity between original and perturbed inputs using embedding similarity, lexical overlap, and retrieval stability.
$\Delta$ Rank denotes the mean absolute rank shift, and O@8 denotes top-8 retrieval overlap.}
\label{tab:similarity}
\resizebox{0.95\columnwidth}{!}{%
\begin{tabular*}{\textwidth}{@{\extracolsep{\fill}}llrrrrrrr}
\toprule
\multirow{2}{*}{\textbf{Dataset}}
& \multirow{2}{*}{\textbf{Split}}
& \multirow{2}{*}{\shortstack{\textbf{Cosine}\\\textbf{sim.}}}
& \multicolumn{2}{c}{\textbf{Lexical overlap}}
& \multicolumn{2}{c}{\textbf{BM25}}
& \multicolumn{2}{c}{\textbf{TF-IDF}} \\
\cmidrule(lr){4-5}
\cmidrule(lr){6-7}
\cmidrule(lr){8-9}
& & 
& \textbf{Jaccard}
& \textbf{Trigram}
& \textbf{$\Delta$ Rank}
& \textbf{O@8}
& \textbf{$\Delta$ Rank}
& \textbf{O@8} \\
\midrule
SST-2
&  & 0.919 & 0.872 & 0.711 & 3.33 & 0.930 & 2.94 & 0.875 \\

\multirow{3}{*}{ProverQA}
& easy & 0.965 & 0.943 & 0.842 & 13.91 & 0.832 & 16.96 & 0.815 \\
& medium & 0.954 & 0.934 & 0.809 & 9.72 & 0.845 & 15.52 & 0.774 \\
& hard & 0.957 & 0.936 & 0.795 & 8.19 & 0.850 & 16.55 & 0.736 \\

\multirow{2}{*}{Problemathic}
& simple & 0.859 & 0.683 & 0.537 & 75.94 & 0.816 & 72.05 & 0.799 \\
& complex & 0.878 & 0.728 & 0.581 & 26.19 & 0.840 & 27.40 & 0.845 \\
\bottomrule
\end{tabular*}
}
\end{table*}

To verify that task preserving perturbations do not simply replace exemplars with unrelated inputs, we quantify original--perturbed similarity using embedding similarity, lexical overlap, and retrieval stability. Specifically, we report sentence-level cosine similarity~\cite{liu2022makes}, token and trigram overlap~\cite{ji2024submodular}, and BM25/TF-IDF rank stability, with detailed metric definitions provided in Appendix~\ref{app:similarity_metrics}.
As shown in Table~\ref{tab:similarity}, the perturbed inputs remain close to their original counterparts under most metrics. SST-2 and ProverQA preserve high embedding and lexical similarity, while PROBLEMATHIC exhibits larger retrieval rank shifts because added numerical distractors change exemplar ranking more strongly. These results support our controlled setup: the perturbations are not arbitrary replacements, but task-related input changes that can still meaningfully affect ICL behavior.

\subsection{Additional Experiments}

\begin{table*}[ht]
\centering
\caption{Accuracy (\%) on SST-2 under different exemplar replacement strategies. For each model and replacement budget, the lowest accuracy is highlighted in bold and the second lowest is underlined.}
\label{tab:position}
\small
\setlength{\tabcolsep}{4.5pt}
\renewcommand{\arraystretch}{1.08}
\resizebox{0.9\columnwidth}{!}{%
\begin{tabular*}{\textwidth}{@{\extracolsep{\fill}}lccccccccc}
\toprule
\multirow{2}{*}{Method}
& \multicolumn{3}{c}{Llama-2-7B}
& \multicolumn{3}{c}{Llama-3.1-8B}
& \multicolumn{3}{c}{Qwen2.5-3B} \\
\cmidrule(lr){2-4} \cmidrule(lr){5-7} \cmidrule(lr){8-10}
& 16 & 24 & 28
& 16 & 24 & 28
& 16 & 24 & 28 \\
\midrule
random & 94.7 & 88.2 & 72.9
       & \underline{96.9} & \underline{94.9} & \underline{88.6}
       & 98.2 & 95.0 & \textbf{93.1} \\
middle & \underline{94.3} & \textbf{74.8} & \underline{61.0}
       & 98.7 & 95.9 & 91.8
       & \underline{95.7} & \underline{93.5} & 98.2 \\
head   & 95.2 & 81.1 & 70.1
       & \textbf{91.5} & \textbf{83.4} & \textbf{85.2}
       & \textbf{89.0} & \textbf{89.8} & \underline{93.6} \\
tail   & \textbf{92.3} & \underline{77.6} & \textbf{58.1}
       & 98.4 & 97.1 & 92.8
       & 96.2 & 95.4 & 98.9 \\
\bottomrule
\end{tabular*}
}
\end{table*}
\vspace{-2mm}

\paragraph{Positional Effects.} 
\label{exp:position}
We further test whether the position of perturbed exemplars affects ICL robustness by comparing placement policies on SST-2 under three structured policies: \textit{head} perturbs the first $K$ exemplars, \textit{middle} perturbs a centered contiguous block, and \textit{tail} perturbs the last $K$ exemplars nearest to the query. Table~\ref{tab:position} shows that \textsc{Llama-2-7B} is consistently more sensitive to tail perturbations than to random replacement. When 16, 24, and 28 exemplars are replaced, tail placement gives 92.3\%, 77.6\%, and 58.1\% accuracy, compared with 94.7\%, 88.2\%, and 72.9\% under random placement. This suggests that smaller models exhibit strong recency sensitivity, making perturbations near the query disproportionately harmful.

\vspace{-2mm}
\begin{figure}[ht]
    \centering
    \includegraphics[width=0.7\linewidth]{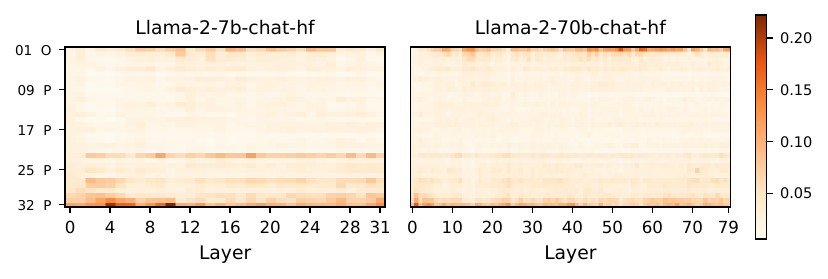}
    \vspace{-2mm}
    \caption{Attention Map under Tail Perturbation.}
    \label{fig:attention}
    \vspace{-2mm}
\end{figure}

\paragraph{Attention Map.}
Motivated by positional effect in Section~\ref{exp:position}, we visualize exemplar level attention under tail perturbations where only the first 4 exemplars are clean out of 32. Figure~\ref{fig:attention} shows that both models attend strongly to recent exemplars near the query but larger models reassign more attention to the early clean exemplars. 
This suggests that larger models better recover task-reliable evidence from mixed contexts, partially explaining their greater robustness to tail perturbations.

%% file: sec/5_conclusion.tex
\section{Conclusion}
\label{sec:conclusion}

This work shows that exemplar correctness is not sufficient for exemplar utility in in-context learning. We introduce task preserving exemplar perturbations, which modify only exemplar inputs while keeping each demonstration valid under the same task mapping. Through the lens of contextual evidence shift, we show how such perturbations can change the effective evidence mixture used for contextual inference and thereby make correct demonstrations harmful. Experiments across sentiment classification, logical reasoning, and math word problems show that task preserving perturbed exemplars can substantially degrade ICL performance. These results suggest that robust ICL should evaluate not only whether demonstrations are correct, but also whether they provide useful contextual evidence for the intended query distribution.
This study also has limitations. Our perturbations are controlled and interpretable, but they cover only part of the broader space of task preserving shifts that may occur in realistic retrieval or prompting pipelines. Our analysis is also limited to text-only ICL with open-weight instruction-tuned models, and the contextual-evidence-shift formulation should be viewed as an explanatory abstraction rather than a complete mechanistic account. Future work should extend this framework to broader perturbation types, multimodal and agentic settings, and defense mechanisms that select or weight demonstrations by utility rather than correctness alone.

%% file: sec/appendix_arxiv.tex
\section*{Appendix}
\vspace{-0.5em}
\hypersetup{pdfborder={0 0 0}}
{
\small
\setlength{\parskip}{0pt}
\startcontents[appendix]
\printcontents[appendix]{}{1}{\setcounter{tocdepth}{2}}
}
\hypersetup{pdfborder={0 0 1}}
\vspace{1em}

\section{Open Science}
\label{app:open_science}
To facilitate reproducibility, we release the code used for prompt construction, exemplar perturbation, model inference, and evaluation. The code repository is available at \repourl.

\section{Theory}
\label{app:theory}

\subsection{Effective Perturbed Evidence Mass}
\label{app:effective_evidence_mass}

In Section~\ref{sec:contextual_evidence_shift}, we interpret task preserving perturbations as shifting the contextual evidence used by ICL. Here we make this interpretation more explicit by distinguishing the nominal perturbation ratio $\rho$ from the effective amount of perturbed evidence used by the model.

Let $D_{\rho}$ denote a perturbed in-context demonstration set with $M$ exemplars, and let $I_{\rho}\subseteq\{1,\ldots,M\}$ be the set of perturbed exemplar indices. For each position $i$, define
\begin{equation}
z_i^{\rho}
=
\begin{cases}
(\tilde{x}_i,\tilde{y}_i), & i\in I_{\rho},\\
(x_i,y_i), & i\notin I_{\rho}.
\end{cases}
\end{equation}
Although the nominal perturbation ratio is $\rho=|I_{\rho}|/M$, under ICL setting a model need not weight all exemplars uniformly. We therefore let
\begin{equation}
w_i
=
w_{\theta}(i\mid x_q,D_{\rho}),
\qquad
w_i\ge 0,
\qquad
\sum_{i=1}^{M}w_i=1,
\end{equation}
where $w_i$ denotes the model's implicit weight on exemplar $i$ when predicting the query $x_q$. This weight can be understood abstractly as the influence assigned to an exemplar, rather than as a direct architectural attention score. It may depend on position, query similarity, surface form, task difficulty, and model scale.

The weighted empirical contextual evidence used by the model can then be written as
\begin{equation}
\widehat{\mathcal{P}}_{\theta}^{\mathcal{T}}(D_{\rho},x_q)
=
\sum_{i=1}^{M}
w_{\theta}(i\mid x_q,D_{\rho})\,
\delta_{z_i^{\rho}}.
\label{eq:weighted-empirical-context}
\end{equation}
We define the effective perturbed evidence mass as
\begin{equation}
m_{\theta}(D_{\rho},x_q)
=
\sum_{i\in I_{\rho}}
w_{\theta}(i\mid x_q,D_{\rho}).
\label{eq:app-effective-evidence-mass}
\end{equation}
This quantity equals the nominal perturbation ratio $\rho$ only when the model weights all exemplars uniformly, i.e., when $w_i=1/M$ for every $i$. In general, $m_{\theta}(D_{\rho},x_q)$ can be larger or smaller than $\rho$. For example, if perturbed exemplars appear near the query and receive disproportionately high weight, then $m_{\theta}>\rho$; if the model discounts perturbed exemplars, then $m_{\theta}<\rho$.

When $0<m_{\theta}<1$, define the normalized weighted empirical distributions over clean and perturbed exemplars as
\begin{equation}
\widehat{\mathcal{P}}^{\mathcal{T}}_{0,\theta}
=
\frac{1}{1-m_{\theta}}
\sum_{i\notin I_{\rho}}
w_i\delta_{(x_i,y_i)},
\label{eq:weighted-clean-context}
\end{equation}
and
\begin{equation}
\widehat{\mathcal{P}}^{\mathcal{T}}_{1,\theta}
=
\frac{1}{m_{\theta}}
\sum_{i\in I_{\rho}}
w_i\delta_{(\tilde{x}_i,\tilde{y}_i)}.
\label{eq:weighted-perturbed-context}
\end{equation}
Then the weighted contextual evidence can be decomposed as the effective mixture
\begin{equation}
\widehat{\mathcal{P}}_{\theta}^{\mathcal{T}}(D_{\rho},x_q)
=
(1-m_{\theta})
\widehat{\mathcal{P}}^{\mathcal{T}}_{0,\theta}
+
m_{\theta}
\widehat{\mathcal{P}}^{\mathcal{T}}_{1,\theta}.
\label{eq:effective-mixture-context}
\end{equation}
Thus, the effective mixture coefficient is $m_{\theta}$ rather than the nominal ratio $\rho$. This distinction explains why the same perturbation budget can have different effects across model scales, exemplar positions, and task difficulties.

This effective-mixture view can also be inserted into the posterior-predictive interpretation from Section~\ref{sec:contextual_evidence_shift}:
\begin{equation}
p_{\theta}(y\mid x_q,D_{\rho})
\approx
\int
p_{\theta}(y\mid x_q,h)
q_{\theta}
\!\left(
h
\mid
(1-m_{\theta})
\widehat{\mathcal{P}}^{\mathcal{T}}_{0,\theta}
+
m_{\theta}
\widehat{\mathcal{P}}^{\mathcal{T}}_{1,\theta}
\right)
dh.
\label{eq:effective-posterior-predictive}
\end{equation}
This equation makes explicit that perturbations influence prediction by changing the contextual hypothesis distribution through the effective perturbed evidence mass. The raw ratio $\rho$ specifies how many exemplars are perturbed, whereas $m_{\theta}$ captures how much perturbed evidence the model effectively uses.

\subsection{Connection to Exemplar Utility}
\label{app:effective_mass_utility}

We next show how the effective perturbed evidence mass connects to the sign of exemplar utility. Let $\mathcal{R}_{Q}(m)$ denote the risk on evaluation distribution $Q$ when the model's effective perturbed evidence mass is $m$. This is an abstraction of the risk $\mathcal{R}_{Q}(f_{\theta}(\cdot\mid D_{\rho}))$ in Section~\ref{sec:contextual_evidence_shift}, where we isolate the dependence on $m_{\theta}$.

Consider a perturbed exemplar $i\in I_{\rho}$ with implicit weight $w_i$. If this exemplar is removed from the context and the remaining weights are renormalized, then the effective perturbed evidence mass changes from $m$ to
\begin{equation}
m_{-i}
=
\frac{m-w_i}{1-w_i}.
\label{eq:remove-perturbed-mass}
\end{equation}
Since $i$ is perturbed, removing it decreases the effective perturbed evidence mass whenever $m<1$. The leave-one-out utility of exemplar $i$ can then be approximated as
\begin{equation}
\Delta \mathcal{R}_{Q}(D,i)
=
\mathcal{R}_{Q}(D\setminus i)
-
\mathcal{R}_{Q}(D)
\approx
\mathcal{R}_{Q}(m_{-i})
-
\mathcal{R}_{Q}(m).
\label{eq:utility-effective-mass}
\end{equation}
Using a first-order Taylor expansion around $m$, we obtain
\begin{equation}
\Delta \mathcal{R}_{Q}(D,i)
\approx
\frac{\partial \mathcal{R}_{Q}(m)}{\partial m}
(m_{-i}-m).
\label{eq:taylor-utility}
\end{equation}
By substituting Eq.~\eqref{eq:remove-perturbed-mass}, we have
\begin{equation}
m_{-i}-m
=
-\frac{w_i(1-m)}{1-w_i}.
\end{equation}
Therefore,
\begin{equation}
\Delta \mathcal{R}_{Q}(D,i)
\approx
-
\frac{w_i(1-m)}{1-w_i}
\frac{\partial \mathcal{R}_{Q}(m)}{\partial m}.
\label{eq:perturbed-utility-sign}
\end{equation}

Eq.~\eqref{eq:perturbed-utility-sign} gives a simple condition for when a task correct perturbed exemplar has negative utility. For clean evaluation queries, let $Q=\mathcal{P}^{\mathcal{T}}_0$. If increasing the effective perturbed evidence mass raises clean test risk,
\begin{equation}
\frac{\partial \mathcal{R}_{\mathcal{P}^{\mathcal{T}}_0}(m)}{\partial m}>0,
\end{equation}
then Eq.~\eqref{eq:perturbed-utility-sign} implies
\begin{equation}
\Delta \mathcal{R}_{\mathcal{P}^{\mathcal{T}}_0}(D,i)<0.
\end{equation}
Therefore, the perturbed exemplar is harmful for clean queries, even though it remains correct under the task mapping $g_{\mathcal{T}}$.

Conversely, for matched perturbed evaluation queries, let $Q=\mathcal{P}^{\mathcal{T}}_1$. If increasing the effective perturbed evidence mass helps the model adapt to the perturbed input regime,
\begin{equation}
\frac{\partial \mathcal{R}_{\mathcal{P}^{\mathcal{T}}_1}(m)}{\partial m}<0,
\end{equation}
then the same perturbed exemplar can have positive utility:
\begin{equation}
\Delta \mathcal{R}_{\mathcal{P}^{\mathcal{T}}_1}(D,i)>0.
\end{equation}
This formalizes the central correctness-utility gap. A perturbed exemplar can satisfy the task mapping and therefore be correct, while still having negative utility on clean queries if it shifts the effective contextual evidence away from the clean evaluation regime. The same exemplar can become neutral or useful when the evaluation distribution matches the perturbed regime.

For completeness, if $i\notin I_{\rho}$ is a clean exemplar, removing it increases the effective perturbed evidence mass:
\begin{equation}
m_{-i}
=
\frac{m}{1-w_i},
\qquad
m_{-i}-m
=
\frac{mw_i}{1-w_i}.
\end{equation}
Hence, when clean test risk increases with $m$, clean exemplars tend to have positive utility on clean queries because their presence reduces the relative mass of perturbed contextual evidence. This provides a complementary interpretation of why exemplar position and attention allocation can modulate perturbation strength: they change the effective mass $m_{\theta}$, not merely the nominal replacement ratio $\rho$.
\section{Experimental Setup}
\label{app:experimental_setup}

\subsection{Exemplar Formatting}
\label{app:exemplar_format}
For all tasks, each in-context exemplar is formatted as an input block followed by its corresponding target output. Below we show simplified examples for sentiment analysis, logical reasoning tasks, and math word problems.

\subsubsection*{Sentiment Analysis}
For sentiment analysis, each exemplar consists of a sentence followed by its sentiment label.

\begin{quote}
\small
sentence: show us a good time \\
Instruction: You are doing sentiment analysis. Only output positive or negative. \\
The answer is positive. \\

sentence: as dumb and cheesy \\
Instruction: You are doing sentiment analysis. Only output positive or negative. \\
The answer is negative. \\

sentence: it’s a charming and often affecting journey \\
Instruction: You are doing sentiment analysis. Only output positive or negative. \\
The answer is
\end{quote}

\subsubsection*{Logical Reasoning Task}
For logical reasoning, each in-context exemplar contains a fact set, a question, candidate options, and a structured target with both the reasoning process and the final answer.

\begin{quote}
\small
Given a problem statement as contexts, the task is to answer a logical reasoning question. Your answer should be in JSON format with keys: reasoning, answer.

Given the facts below, answer the question. \\

Facts: Alice is diligent. If someone is diligent, then they finish their work. \\

Question: Based on the above information, is the following statement true, false, or uncertain? Alice finishes her work. \\

Options: ["A) True", "B) False", "C) Uncertain"] \\

\{ \\
\hspace*{1em}"reasoning": "fact1: Alice is diligent.\textbackslash nrule: If someone is diligent, then they finish their work.\textbackslash nconclusion: Alice finishes her work.\textbackslash n\textbackslash nTherefore, it is true that Alice finishes her work. The correct option is: A.", \\
\hspace*{1em}"answer": "A" \\
\} \\[0.5em]

Given the facts below, answer the question. \\

Facts: Carol is careful. If someone is careful, then they avoid mistakes. \\

Question: Based on the above information, is the following statement true, false, or uncertain? Carol avoids mistakes. \\

Options: ["A) True", "B) False", "C) Uncertain"] \\
\end{quote}

\subsubsection*{Math Word Problems}
For math word problems, each in-context exemplar contains a passage, a question, and a structured target with both the reasoning process and the final numeric answer.

\begin{quote}
\small
Solve the math word problem. Your answer should be in JSON format with keys: reasoning, answer. The answer value should be numeric.

\textless\textless Passage\textgreater\textgreater There are 33 oak trees currently in the park. Park workers had to cut down 18 oak trees that were damaged.
\textless\textless Question\textgreater\textgreater How many oak trees will be in the park when the workers are finished? \\

\{ \\
\hspace*{1em}"reasoning": "Explanation: To find the number of oak trees remaining in the park, subtract the number of trees cut down from the initial number of trees. Thus, 33 - 18 = 15.", \\
\hspace*{1em}"answer": "15.0" \\
\} \\[0.5em]

\textless\textless Passage\textgreater\textgreater The town of Milburg has 5256 grown-ups and 2987 children.
\textless\textless Question\textgreater\textgreater How many people live in Milburg?
\end{quote}

\subsection{Perturbation Instantiations}
\label{app:perturbation_details}

Table~\ref{tab:perturbation_instantiations} summarizes how each dataset instantiates the perturbation regimes defined in Section~\ref{sec:method}.

\begin{table}[h]
\centering
\small
\setlength{\tabcolsep}{6pt}
\renewcommand{\arraystretch}{1.08}
\caption{Dataset Specific Perturbation Instantiations.}
\label{tab:perturbation_instantiations}
\begin{tabular}{llll}
\toprule
Dataset & Regime & Perturbation type & Target relation \\
\midrule
SST-2 
& Task preserving 
& Sentiment changing input edits 
& $\tilde{y}_i = g_{\mathcal{T}}(\tilde{x}_i)$\\

ProverQA 
& Label preserving 
& Distractor insertion and premise reordering 
& $\tilde{y}_i = y_i$ \\

PROBLEMATHIC 
& Answer preserving 
& Irrelevant numerical information 
& $\tilde{y}_i = y_i$ \\
\bottomrule
\end{tabular}
\end{table}

\subsection{Similarity Metric Definitions}
\label{app:similarity_metrics}

We provide the detailed definitions of the similarity and retrieval-stability metrics used in Section~\ref{sec:similarity}. For each exemplar pair, let $x_i$ denote the original input and $\tilde{x}_i$ denote its perturbed counterpart.

\paragraph{Embedding similarity.}
We encode each input with a RoBERTa encoder and obtain a sentence-level representation by mean pooling over the last-layer token embeddings. Let $h_{it}$ be the last-layer hidden state of token $t$ in $x_i$, and let $m_{it}$ be its attention mask. The sentence embedding is computed as
\[
e(x_i) =
\frac{\sum_{t} m_{it} h_{it}}{\sum_{t} m_{it}} .
\]
The embedding similarity between the original and perturbed inputs is then measured by cosine similarity:
\[
\mathrm{CosSim}(x_i,\tilde{x}_i)
=
\frac{e(x_i)^\top e(\tilde{x}_i)}
{\|e(x_i)\|_2 \|e(\tilde{x}_i)\|_2}.
\]
For each dataset split, we report the average cosine similarity over all original--perturbed exemplar pairs.

\paragraph{Lexical overlap.}
We tokenize each input using a regular-expression tokenizer that separates word tokens and punctuation. Let $\mathcal{T}(x_i)$ and $\mathcal{T}(\tilde{x}_i)$ denote the token sets of the original and perturbed inputs. Token-level Jaccard overlap is defined as
\[
\mathrm{Jaccard}(x_i,\tilde{x}_i)
=
\frac{
|\mathcal{T}(x_i) \cap \mathcal{T}(\tilde{x}_i)|
}{
|\mathcal{T}(x_i) \cup \mathcal{T}(\tilde{x}_i)|
}.
\]
We also compute $n$-gram overlap. Let $\mathcal{G}_n(x_i)$ denote the set of contiguous token $n$-grams in $x_i$. The $n$-gram overlap is
\[
\mathrm{Overlap}_n(x_i,\tilde{x}_i)
=
\frac{
|\mathcal{G}_n(x_i) \cap \mathcal{G}_n(\tilde{x}_i)|
}{
|\mathcal{G}_n(x_i) \cup \mathcal{G}_n(\tilde{x}_i)|
}.
\]
In the main table, we report trigram overlap, i.e., $n=3$, as a stricter surface-form similarity measure.

\paragraph{Retrieval stability.}
To evaluate whether perturbations change retrieval based exemplar selection, we compare rankings induced by the original exemplar pool and the perturbed exemplar pool. For each test query $q_j$, we rank all original exemplars $\{x_i\}_{i=1}^{M}$ and all perturbed exemplars $\{\tilde{x}_i\}_{i=1}^{M}$ using either BM25 or TF-IDF cosine similarity. Let
\[
r_{ij}
\]
be the rank of original exemplar $x_i$ for query $q_j$, and let
\[
\tilde{r}_{ij}
\]
be the rank of its perturbed counterpart $\tilde{x}_i$ for the same query. We define the rank shift as
\[
\Delta r_{ij} = \tilde{r}_{ij} - r_{ij}.
\]
The main table reports the mean absolute rank shift:
\[
\Delta r_{\mathrm{abs}}
=
\frac{1}{NM}
\sum_{j=1}^{N}
\sum_{i=1}^{M}
|\Delta r_{ij}|,
\]
where $N$ is the number of test queries and $M$ is the number of exemplars.

We also report top-$k$ retrieval overlap. Let $\mathcal{R}_{j}^{k}$ be the set of exemplar indices in the top-$k$ results from the original exemplar pool for query $q_j$, and let $\tilde{\mathcal{R}}_{j}^{k}$ be the corresponding top-$k$ set from the perturbed exemplar pool. The overlap is defined as
\[
\mathrm{Overlap@}k
=
\frac{1}{N}
\sum_{j=1}^{N}
\frac{
|\mathcal{R}_{j}^{k} \cap \tilde{\mathcal{R}}_{j}^{k}|
}{k}.
\]
In the main table, we report $\mathrm{Overlap@}8$.

\subsection{Computational Resource}
\label{app:computational_resource}
All experiments are conducted using 2 NVIDIA RTX PRO 6000 GPUs
with 96 GB of memory and a 64-core AMD EPYC™ 9554 CPU operating at 3.10GHz. Our codebase is implemented in PyTorch and built on the Hugging Face Transformers library for experimental evaluation, with vLLM~\cite{kwon2023efficient} additionally used to accelerate inference.

\section{Reproducibility}
\label{app:reproducibility}

\subsection{Model and Inference Configuration}
\begin{table*}[ht]
\centering
\caption{Model and core inference settings.}
\label{app:tab:model_inference_details}
\small
\setlength{\tabcolsep}{5pt}
\renewcommand{\arraystretch}{1.12}
\resizebox{\textwidth}{!}{%
\begin{tabular}{llccc}
\toprule
Model family & Model IDs & Backend & Dtype & Decoding \\
\midrule
Llama-2 Chat
&
\texttt{meta-llama/Llama-2-\{7b,13b,70b\}-chat-hf}
&
vLLM
&
bfloat16
&
temperature $=0.0$, top-$p=1.0$
\\

Llama-3.1 Instruct
&
\texttt{meta-llama/Llama-3.1-\{8B,70B\}-Instruct}
&
vLLM
&
bfloat16
&
temperature $=0.0$, top-$p=1.0$
\\

Qwen2.5 Instruct
&
\texttt{Qwen/Qwen2.5-\{3B,7B,14B,32B,72B\}-Instruct}
&
vLLM
&
bfloat16
&
temperature $=0.0$, top-$p=1.0$
\\

Gemma-4 IT
&
\texttt{google/gemma-4-\{E2B,E4B,31B\}-it}
&
vLLM
&
bfloat16
&
temperature $=0.0$, top-$p=1.0$
\\

Qwen3.5
&
\texttt{Qwen/Qwen3.5-\{2B,9B,27B\}}
&
vLLM
&
bfloat16
&
temperature $=0.0$, top-$p=1.0$
\\
\bottomrule
\end{tabular}}
\end{table*}

\paragraph{Hyperparameters.}
As shown in Table~\ref{app:tab:model_inference_details}, all models are evaluated with vLLM using bfloat16 precision, temperature $0.0$, and the default top-$p$ value of $1.0$. For classification tasks, we generate a single token and restrict decoding to the valid label tokens. For ProverQA, we allow up to 2048 new tokens, while for math word problems we allow up to 256 new tokens.

\paragraph{Prompt Construction.} 
All experiments follow the implementation in our evaluation scripts. Llama-2 models use raw ICL prompts without the Hugging Face chat template. Llama-3.1 and Qwen2.5 models use the Hugging Face tokenizer chat template. Gemma-4 and Qwen3.5 models are evaluated through \texttt{icl\_vlm.py}; when a processor chat template is available, we use it in text-only mode, otherwise we fall back to the tokenizer chat template. For Qwen3.5 models, we set \texttt{enable\_thinking=False} whenever this option is supported by the tokenizer or processor.

\noindent\textbf{Answer Parsing.}
For classification, predicted and gold labels are stripped and compared case-insensitively against the configured label set. For ProverQA, we first parse fenced JSON, then any JSON object span, and finally the full output; only the \texttt{answer} field is used for scoring, with a fallback to the first label-like character such as A/B/C. For math, we first parse the JSON \texttt{answer}; if JSON parsing fails, we search for \texttt{\#\#\#\# <number>}, explicit final-answer patterns, and then the last numeric span. Numeric answers are normalized by removing commas and accepting fractions, and correctness uses absolute or relative tolerance $10^{-6}$.

\subsection{Model Details}
\label{app:model_details}
We evaluate a diverse set of open-weight large language models to study how model scale and model family affect robustness to task preserving exemplar perturbations. Specifically, we use the following instruction-tuned model families:

\begin{itemize}
    \item \textbf{Llama-2:} https://huggingface.co/collections/meta-llama/llama-2-family
    \item \textbf{Llama-3.1:} https://huggingface.co/collections/meta-llama/llama-31
    \item \textbf{Qwen-2.5:} https://huggingface.co/collections/Qwen/qwen25
    \item \textbf{Qwen-3.5:} https://huggingface.co/collections/Qwen/qwen35
    \item \textbf{Gemma-4:} https://huggingface.co/collections/google/gemma-4
\end{itemize}

\subsection{Dataset Details}
\label{app:dataset_details}

We evaluate task preserving exemplar perturbations on three datasets covering sentiment classification, logical reasoning, and math word problems. Specifically, we use the following datasets:

\begin{itemize}
    \item \textbf{AdvGLUE:} https://huggingface.co/datasets/AI-Secure/adv\_glue
    \item \textbf{ProverQA:} https://huggingface.co/datasets/opendatalab/ProverQA
    \item \textbf{PROBLEMATHIC:} https://huggingface.co/datasets/him1411/problemathic
\end{itemize}

\section{Additional Experiments}

\subsection{Format Similarity Ablation.}
\begin{figure}[ht]
    \centering
    \includegraphics[width=\linewidth]{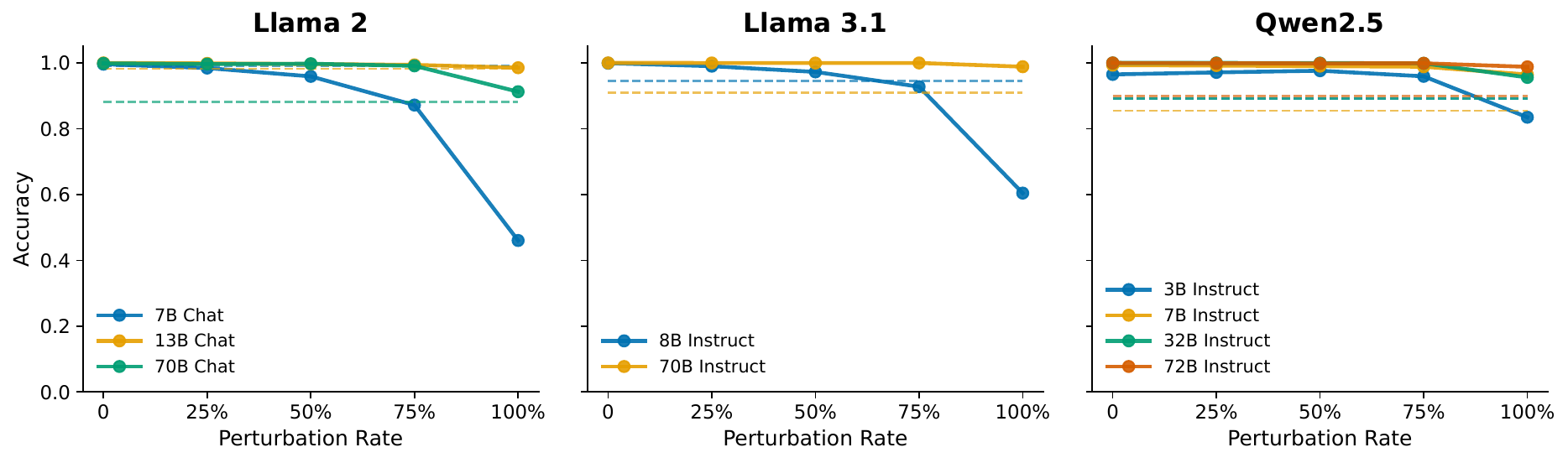}
    \caption{SST-2 accuracy under reduced exemplar-format similarity.}
    \label{fig:sst2_structure}
\end{figure}

We further test whether the surface-form similarity between original and perturbed exemplars affects ICL robustness. Unlike the main SST-2 setting where perturbations are nearly format-preserving, we construct a reduced-similarity variant where the original input \textit{``this film is wonderful''} is paired with the same task preserving adversarial alternative, \textit{``I can't say, given all that I've seen over the years, that this film is wonderful.''}. As shown in Figure~\ref{fig:sst2_structure}, reducing format similarity leads to milder degradation at low perturbation ratios, indicating that highly similar input perturbations are more effective at reducing ICL accuracy. This suggests that the vulnerability is not only caused by the presence of perturbed exemplars, but also by how closely they mimic the original exemplar form: adversarial inputs that preserve the original surface structure can be harder for models to discount and therefore induce stronger performance drops.

\section{Additional Results}
\label{app:additional_results}

\subsection{All Results in Sentiment Analysis}
\label{app:all_sentiment}

\begin{figure}[ht]
    \centering
    \includegraphics[width=\linewidth]{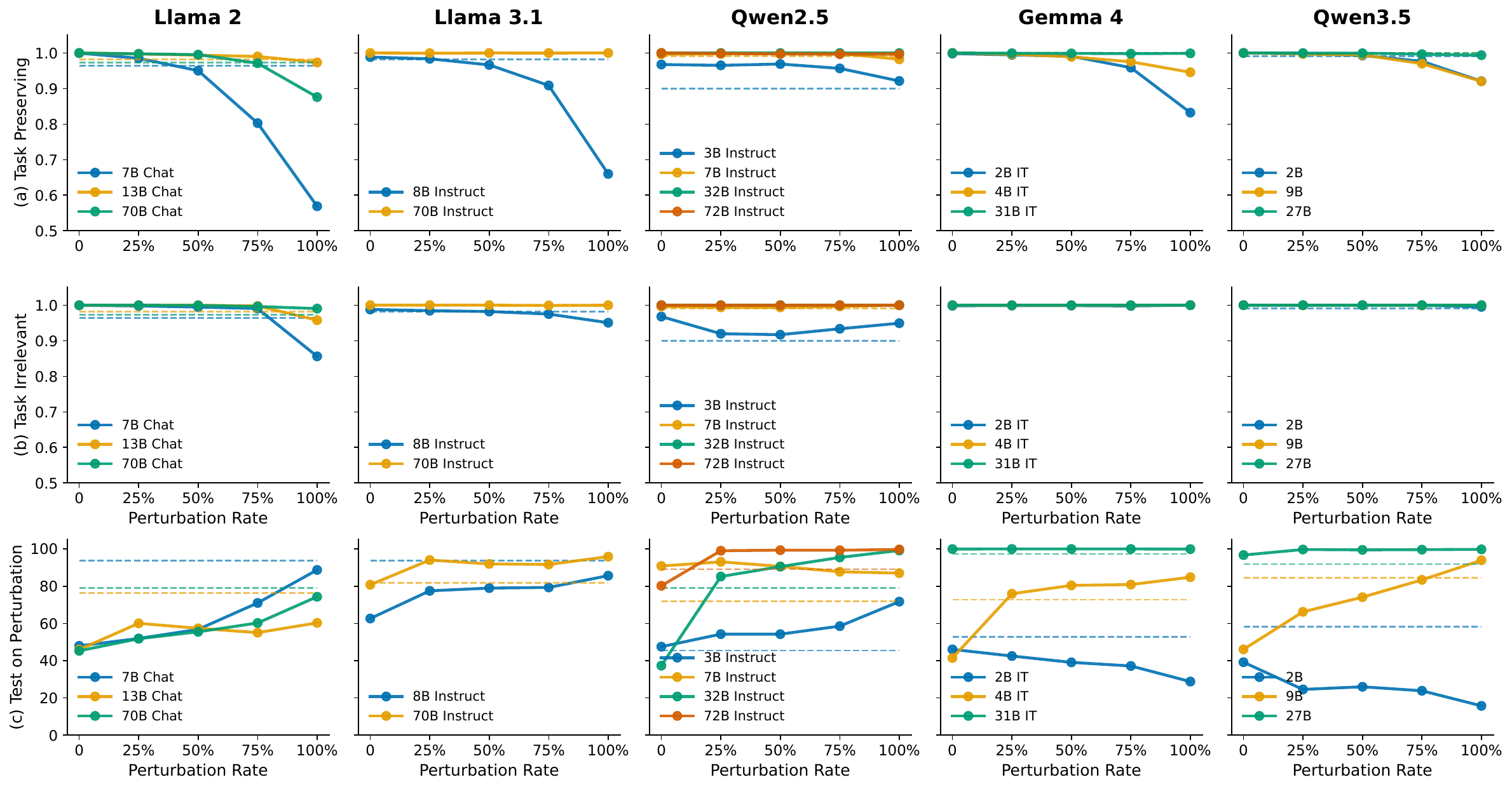}
    \caption{Complete SST-2 sentiment analysis results.}
    \label{fig:sst2_all_results}
\end{figure}

Figure~\ref{fig:sst2_all_results} reports the complete SST-2 results across all evaluated model families. 
We include three complementary evaluations. 
Panel~(a) shows the main task preserving perturbation setting, where selected in-context exemplars are replaced by semantically edited counterparts and paired with task-correct labels under the same sentiment mapping. 
Panel~(b) reports a task irrelevant control, where exemplar inputs are replaced by sentiment-neutral factual statements while the prompt format and output space are kept fixed. 
Panel~(c) evaluates the same perturbed prompts on the matched perturbed test set.

Table~\ref{app:tab:sst2_original_results},~\ref{app:tab:sst2_irrelevant_results} and~\ref{app:tab:sst2_perturbed_results} report the exact numerical results corresponding to Figure~\ref{fig:sst2_all_results}. 
All values are reported as mean accuracy $\pm$ standard deviation over 100 random runs, in percentage points.

\begin{table*}[ht]
\centering
\scriptsize
\setlength{\tabcolsep}{3.0pt}
\caption{\textbf{Complete SST-2 results on the original test set.} 
Accuracy is reported as mean $\pm$ standard deviation over 100 random runs, in percentage points. 
Selected in-context exemplars are replaced by task preserving perturbed exemplars, while evaluation is conducted on the original SST-2 test set.}
\label{app:tab:sst2_original_results}
\resizebox{\linewidth}{!}{
\begin{tabular}{llcccccc}
\toprule
Family & Model & Zero-shot & 0\% & 25\% & 50\% & 75\% & 100\% \\
\midrule
LLAMA-2 & 7B Chat 
& $96.4{\scriptsize\pm0.0}$ & $100.0{\scriptsize\pm0.2}$ & $98.5{\scriptsize\pm2.3}$ & $95.0{\scriptsize\pm3.7}$ & $80.3{\scriptsize\pm8.5}$ & $56.9{\scriptsize\pm7.4}$ \\
LLAMA-2 & 13B Chat 
& $98.2{\scriptsize\pm0.0}$ & $100.0{\scriptsize\pm0.1}$ & $99.7{\scriptsize\pm0.6}$ & $99.4{\scriptsize\pm0.9}$ & $99.0{\scriptsize\pm1.2}$ & $97.3{\scriptsize\pm2.2}$ \\
LLAMA-2 & 70B Chat 
& $97.3{\scriptsize\pm0.0}$ & $100.0{\scriptsize\pm0.0}$ & $99.8{\scriptsize\pm0.9}$ & $99.5{\scriptsize\pm1.1}$ & $97.1{\scriptsize\pm4.0}$ & $87.6{\scriptsize\pm11.8}$ \\
LLAMA-3.1 & 8B Instruct 
& $98.2{\scriptsize\pm0.0}$ & $98.9{\scriptsize\pm2.7}$ & $98.4{\scriptsize\pm3.1}$ & $96.6{\scriptsize\pm3.5}$ & $90.9{\scriptsize\pm7.7}$ & $66.0{\scriptsize\pm12.2}$ \\
LLAMA-3.1 & 70B Instruct 
& $100.0{\scriptsize\pm0.0}$ & $100.0{\scriptsize\pm0.0}$ & $99.9{\scriptsize\pm0.2}$ & $100.0{\scriptsize\pm0.1}$ & $100.0{\scriptsize\pm0.1}$ & $100.0{\scriptsize\pm0.0}$ \\
QWEN2.5 & 3B Instruct 
& $90.0{\scriptsize\pm0.0}$ & $96.8{\scriptsize\pm3.0}$ & $96.5{\scriptsize\pm4.0}$ & $96.9{\scriptsize\pm3.4}$ & $95.7{\scriptsize\pm6.1}$ & $92.1{\scriptsize\pm5.6}$ \\
QWEN2.5 & 7B Instruct 
& $99.1{\scriptsize\pm0.0}$ & $99.6{\scriptsize\pm0.4}$ & $99.7{\scriptsize\pm0.4}$ & $99.8{\scriptsize\pm0.4}$ & $99.9{\scriptsize\pm0.3}$ & $98.2{\scriptsize\pm2.6}$ \\
QWEN2.5 & 32B Instruct 
& $100.0{\scriptsize\pm0.0}$ & $100.0{\scriptsize\pm0.0}$ & $100.0{\scriptsize\pm0.0}$ & $100.0{\scriptsize\pm0.1}$ & $100.0{\scriptsize\pm0.1}$ & $100.0{\scriptsize\pm0.2}$ \\
QWEN2.5 & 72B Instruct 
& $100.0{\scriptsize\pm0.0}$ & $100.0{\scriptsize\pm0.0}$ & $99.9{\scriptsize\pm0.3}$ & $99.7{\scriptsize\pm0.5}$ & $99.7{\scriptsize\pm0.5}$ & $99.6{\scriptsize\pm0.5}$ \\
GEMMA-4 & 2B IT 
& $100.0{\scriptsize\pm0.0}$ & $99.8{\scriptsize\pm0.4}$ & $99.5{\scriptsize\pm0.6}$ & $99.1{\scriptsize\pm0.8}$ & $95.9{\scriptsize\pm2.8}$ & $83.2{\scriptsize\pm4.8}$ \\
GEMMA-4 & 4B IT 
& $100.0{\scriptsize\pm0.0}$ & $100.0{\scriptsize\pm0.0}$ & $99.6{\scriptsize\pm0.6}$ & $98.9{\scriptsize\pm1.2}$ & $97.5{\scriptsize\pm2.0}$ & $94.6{\scriptsize\pm3.9}$ \\
GEMMA-4 & 31B IT 
& $100.0{\scriptsize\pm0.0}$ & $100.0{\scriptsize\pm0.0}$ & $99.9{\scriptsize\pm0.2}$ & $99.9{\scriptsize\pm0.3}$ & $99.8{\scriptsize\pm0.4}$ & $99.9{\scriptsize\pm0.3}$ \\
QWEN3.5 & 2B 
& $99.1{\scriptsize\pm0.0}$ & $100.0{\scriptsize\pm0.0}$ & $99.8{\scriptsize\pm0.4}$ & $99.3{\scriptsize\pm1.2}$ & $97.7{\scriptsize\pm3.8}$ & $92.1{\scriptsize\pm7.6}$ \\
QWEN3.5 & 9B 
& $100.0{\scriptsize\pm0.0}$ & $100.0{\scriptsize\pm0.0}$ & $99.7{\scriptsize\pm0.7}$ & $99.5{\scriptsize\pm1.0}$ & $97.0{\scriptsize\pm5.4}$ & $92.0{\scriptsize\pm9.5}$ \\
QWEN3.5 & 27B 
& $100.0{\scriptsize\pm0.0}$ & $100.0{\scriptsize\pm0.0}$ & $100.0{\scriptsize\pm0.3}$ & $99.9{\scriptsize\pm0.2}$ & $99.7{\scriptsize\pm0.7}$ & $99.4{\scriptsize\pm1.0}$ \\
\bottomrule
\end{tabular}
}
\end{table*}

\begin{table*}[ht]
\centering
\scriptsize
\setlength{\tabcolsep}{3.0pt}
\caption{\textbf{Complete SST-2 results under the task irrelevant control.} 
Accuracy is reported as mean $\pm$ standard deviation over 100 random runs, in percentage points. 
Selected exemplar inputs are replaced with sentiment-neutral factual statements while the prompt format, output space, and evaluation set are kept fixed.}
\label{app:tab:sst2_irrelevant_results}
\resizebox{\linewidth}{!}{
\begin{tabular}{llcccccc}
\toprule
Family & Model & Zero-shot & 0\% & 25\% & 50\% & 75\% & 100\% \\
\midrule
LLAMA-2 & 7B Chat 
& $96.4{\scriptsize\pm0.0}$ & $100.0{\scriptsize\pm0.2}$ & $99.8{\scriptsize\pm0.7}$ & $99.5{\scriptsize\pm1.1}$ & $99.1{\scriptsize\pm1.8}$ & $85.6{\scriptsize\pm3.2}$ \\
LLAMA-2 & 13B Chat 
& $98.2{\scriptsize\pm0.0}$ & $100.0{\scriptsize\pm0.1}$ & $100.0{\scriptsize\pm0.0}$ & $100.0{\scriptsize\pm0.1}$ & $99.8{\scriptsize\pm0.4}$ & $95.8{\scriptsize\pm3.2}$ \\
LLAMA-2 & 70B Chat 
& $97.3{\scriptsize\pm0.0}$ & $100.0{\scriptsize\pm0.0}$ & $100.0{\scriptsize\pm0.1}$ & $99.9{\scriptsize\pm0.3}$ & $99.6{\scriptsize\pm1.4}$ & $99.0{\scriptsize\pm1.1}$ \\
LLAMA-3.1 & 8B Instruct 
& $98.2{\scriptsize\pm0.0}$ & $98.8{\scriptsize\pm0.2}$ & $98.5{\scriptsize\pm0.4}$ & $98.2{\scriptsize\pm0.7}$ & $97.5{\scriptsize\pm1.5}$ & $95.1{\scriptsize\pm1.8}$ \\
LLAMA-3.1 & 70B Instruct 
& $100.0{\scriptsize\pm0.0}$ & $100.0{\scriptsize\pm0.0}$ & $100.0{\scriptsize\pm0.0}$ & $100.0{\scriptsize\pm0.0}$ & $99.9{\scriptsize\pm0.3}$ & $100.0{\scriptsize\pm0.1}$ \\
QWEN2.5 & 3B Instruct 
& $90.0{\scriptsize\pm0.0}$ & $96.8{\scriptsize\pm2.9}$ & $92.0{\scriptsize\pm5.0}$ & $91.7{\scriptsize\pm5.8}$ & $93.4{\scriptsize\pm5.8}$ & $94.9{\scriptsize\pm3.2}$ \\
QWEN2.5 & 7B Instruct 
& $99.1{\scriptsize\pm0.0}$ & $99.6{\scriptsize\pm0.4}$ & $99.4{\scriptsize\pm0.4}$ & $99.4{\scriptsize\pm0.4}$ & $99.6{\scriptsize\pm0.4}$ & $100.0{\scriptsize\pm0.0}$ \\
QWEN2.5 & 32B Instruct 
& $100.0{\scriptsize\pm0.0}$ & $100.0{\scriptsize\pm0.0}$ & $100.0{\scriptsize\pm0.0}$ & $100.0{\scriptsize\pm0.0}$ & $100.0{\scriptsize\pm0.1}$ & $100.0{\scriptsize\pm0.0}$ \\
QWEN2.5 & 72B Instruct 
& $100.0{\scriptsize\pm0.0}$ & $100.0{\scriptsize\pm0.0}$ & $100.0{\scriptsize\pm0.0}$ & $100.0{\scriptsize\pm0.0}$ & $100.0{\scriptsize\pm0.0}$ & $100.0{\scriptsize\pm0.0}$ \\
GEMMA-4 & 2B IT 
& $100.0{\scriptsize\pm0.0}$ & $99.8{\scriptsize\pm0.3}$ & $99.9{\scriptsize\pm0.3}$ & $99.9{\scriptsize\pm0.3}$ & $99.8{\scriptsize\pm0.5}$ & $100.0{\scriptsize\pm0.0}$ \\
GEMMA-4 & 4B IT 
& $100.0{\scriptsize\pm0.0}$ & $100.0{\scriptsize\pm0.0}$ & $100.0{\scriptsize\pm0.2}$ & $100.0{\scriptsize\pm0.1}$ & $100.0{\scriptsize\pm0.2}$ & $100.0{\scriptsize\pm0.0}$ \\
GEMMA-4 & 31B IT 
& $100.0{\scriptsize\pm0.0}$ & $100.0{\scriptsize\pm0.0}$ & $100.0{\scriptsize\pm0.1}$ & $100.0{\scriptsize\pm0.0}$ & $100.0{\scriptsize\pm0.0}$ & $100.0{\scriptsize\pm0.0}$ \\
QWEN3.5 & 2B 
& $99.1{\scriptsize\pm0.0}$ & $100.0{\scriptsize\pm0.0}$ & $100.0{\scriptsize\pm0.0}$ & $100.0{\scriptsize\pm0.0}$ & $100.0{\scriptsize\pm0.1}$ & $99.5{\scriptsize\pm0.9}$ \\
QWEN3.5 & 9B 
& $100.0{\scriptsize\pm0.0}$ & $100.0{\scriptsize\pm0.0}$ & $100.0{\scriptsize\pm0.2}$ & $100.0{\scriptsize\pm0.2}$ & $99.9{\scriptsize\pm0.4}$ & $100.0{\scriptsize\pm0.0}$ \\
QWEN3.5 & 27B 
& $100.0{\scriptsize\pm0.0}$ & $100.0{\scriptsize\pm0.0}$ & $100.0{\scriptsize\pm0.0}$ & $100.0{\scriptsize\pm0.0}$ & $100.0{\scriptsize\pm0.0}$ & $100.0{\scriptsize\pm0.0}$ \\
\bottomrule
\end{tabular}
}
\end{table*}

\begin{table*}[ht]
\centering
\scriptsize
\setlength{\tabcolsep}{3.0pt}
\caption{\textbf{Complete SST-2 results on the matched perturbed test set.} 
Accuracy is reported as mean $\pm$ standard deviation over 100 random runs, in percentage points. 
The same perturbed prompts are evaluated on perturbed test inputs, allowing us to test whether perturbed exemplars become more useful when the contextual evidence and evaluation distribution are matched.}
\label{app:tab:sst2_perturbed_results}
\resizebox{\linewidth}{!}{
\begin{tabular}{llcccccc}
\toprule
Family & Model & Zero-shot & 0\% & 25\% & 50\% & 75\% & 100\% \\
\midrule
LLAMA-2 & 7B Chat 
& $93.6{\scriptsize\pm0.0}$ & $48.0{\scriptsize\pm2.6}$ & $51.9{\scriptsize\pm8.9}$ & $56.8{\scriptsize\pm9.3}$ & $71.0{\scriptsize\pm11.3}$ & $88.7{\scriptsize\pm6.6}$ \\
LLAMA-2 & 13B Chat 
& $76.4{\scriptsize\pm0.0}$ & $46.0{\scriptsize\pm0.8}$ & $60.1{\scriptsize\pm11.2}$ & $57.4{\scriptsize\pm12.7}$ & $55.1{\scriptsize\pm11.7}$ & $60.3{\scriptsize\pm8.8}$ \\
LLAMA-2 & 70B Chat 
& $79.1{\scriptsize\pm0.0}$ & $45.3{\scriptsize\pm0.5}$ & $51.8{\scriptsize\pm8.4}$ & $55.5{\scriptsize\pm9.6}$ & $60.2{\scriptsize\pm9.9}$ & $74.3{\scriptsize\pm11.3}$ \\
LLAMA-3.1 & 8B Instruct 
& $93.6{\scriptsize\pm0.0}$ & $62.6{\scriptsize\pm9.3}$ & $77.5{\scriptsize\pm9.7}$ & $79.0{\scriptsize\pm8.7}$ & $79.3{\scriptsize\pm8.1}$ & $85.6{\scriptsize\pm7.2}$ \\
LLAMA-3.1 & 70B Instruct 
& $81.8{\scriptsize\pm0.0}$ & $80.8{\scriptsize\pm4.8}$ & $94.0{\scriptsize\pm5.9}$ & $91.9{\scriptsize\pm5.2}$ & $91.6{\scriptsize\pm5.3}$ & $95.8{\scriptsize\pm3.7}$ \\
QWEN2.5 & 3B Instruct 
& $45.5{\scriptsize\pm0.0}$ & $47.5{\scriptsize\pm2.6}$ & $54.3{\scriptsize\pm10.3}$ & $54.2{\scriptsize\pm8.6}$ & $58.5{\scriptsize\pm10.0}$ & $71.7{\scriptsize\pm9.7}$ \\
QWEN2.5 & 7B Instruct 
& $71.8{\scriptsize\pm0.0}$ & $90.8{\scriptsize\pm5.1}$ & $93.1{\scriptsize\pm4.8}$ & $90.6{\scriptsize\pm5.6}$ & $87.7{\scriptsize\pm6.5}$ & $87.0{\scriptsize\pm5.6}$ \\
QWEN2.5 & 32B Instruct 
& $79.1{\scriptsize\pm0.0}$ & $37.3{\scriptsize\pm4.6}$ & $85.2{\scriptsize\pm7.7}$ & $90.5{\scriptsize\pm5.6}$ & $95.4{\scriptsize\pm3.3}$ & $99.2{\scriptsize\pm0.9}$ \\
QWEN2.5 & 72B Instruct 
& $89.1{\scriptsize\pm0.0}$ & $80.2{\scriptsize\pm5.8}$ & $99.0{\scriptsize\pm1.4}$ & $99.3{\scriptsize\pm1.1}$ & $99.3{\scriptsize\pm1.2}$ & $99.7{\scriptsize\pm0.5}$ \\
GEMMA-4 & 2B IT 
& $52.7{\scriptsize\pm0.0}$ & $46.1{\scriptsize\pm0.6}$ & $42.5{\scriptsize\pm5.9}$ & $39.1{\scriptsize\pm6.6}$ & $37.2{\scriptsize\pm8.4}$ & $28.8{\scriptsize\pm10.1}$ \\
GEMMA-4 & 4B IT 
& $72.7{\scriptsize\pm0.0}$ & $41.4{\scriptsize\pm8.3}$ & $75.9{\scriptsize\pm10.0}$ & $80.4{\scriptsize\pm7.2}$ & $80.9{\scriptsize\pm5.4}$ & $84.8{\scriptsize\pm4.7}$ \\
GEMMA-4 & 31B IT 
& $97.3{\scriptsize\pm0.0}$ & $99.9{\scriptsize\pm0.3}$ & $100.0{\scriptsize\pm0.1}$ & $100.0{\scriptsize\pm0.2}$ & $100.0{\scriptsize\pm0.3}$ & $99.9{\scriptsize\pm0.4}$ \\
QWEN3.5 & 2B 
& $58.2{\scriptsize\pm0.0}$ & $39.2{\scriptsize\pm6.7}$ & $24.6{\scriptsize\pm9.6}$ & $26.0{\scriptsize\pm10.0}$ & $23.8{\scriptsize\pm8.8}$ & $15.8{\scriptsize\pm7.3}$ \\
QWEN3.5 & 9B 
& $84.5{\scriptsize\pm0.0}$ & $46.1{\scriptsize\pm8.3}$ & $66.2{\scriptsize\pm12.4}$ & $74.1{\scriptsize\pm8.3}$ & $83.4{\scriptsize\pm7.4}$ & $93.9{\scriptsize\pm4.1}$ \\
QWEN3.5 & 27B 
& $91.8{\scriptsize\pm0.0}$ & $96.7{\scriptsize\pm1.9}$ & $99.7{\scriptsize\pm0.8}$ & $99.5{\scriptsize\pm0.9}$ & $99.6{\scriptsize\pm0.9}$ & $99.8{\scriptsize\pm0.6}$ \\
\bottomrule
\end{tabular}
}
\end{table*}
\clearpage

\subsection{All Results in Logic Reasoning Problems}
\label{app:all_qa}

\providecommand{\accstd}[2]{#1{\scriptsize$\pm$#2}}

Tables~\ref{app:tab:proverqa_easy}--\ref{app:tab:proverqa_hard}
report the complete ProverQA results across all evaluated models and difficulty levels.
Accuracy is reported as mean $\pm$ standard deviation over 5 random runs, in percentage points.
Since each prompt contains 8 in-context exemplars, perturbing 2, 4, 6, and 8 exemplars corresponds
to perturbation ratios of 25\%, 50\%, 75\%, and 100\%, respectively.

\begin{table*}[ht]
\centering
\scriptsize
\setlength{\tabcolsep}{3.2pt}
\renewcommand{\arraystretch}{1.08}
\caption{\textbf{Complete ProverQA results on the Easy split.}
Accuracy is reported as mean $\pm$ standard deviation over 5 random runs.}
\label{app:tab:proverqa_easy}
\resizebox{\textwidth}{!}{%
\begin{tabular}{llcccccc}
\toprule
\textbf{Family} & \textbf{Model} & \textbf{Zero-shot} & \textbf{0\%} & \textbf{25\%} & \textbf{50\%} & \textbf{75\%} & \textbf{100\%} \\
\midrule
Llama-3.1 & 8B Instruct  & \accstd{71.0}{0.0} & \accstd{79.7}{3.2} & \accstd{80.9}{3.4} & \accstd{80.6}{3.4} & \accstd{80.3}{2.0} & \accstd{80.1}{3.6} \\
Llama-3.1 & 70B Instruct & \accstd{92.0}{0.0} & \accstd{96.1}{1.5} & \accstd{96.6}{0.8} & \accstd{96.9}{0.6} & \accstd{96.6}{1.2} & \accstd{96.4}{1.2} \\
Qwen2.5 & 7B Instruct   & \accstd{83.0}{0.0} & \accstd{87.7}{1.5} & \accstd{88.4}{1.5} & \accstd{87.8}{0.7} & \accstd{88.4}{0.8} & \accstd{88.2}{0.7} \\
Qwen2.5 & 14B Instruct  & \accstd{92.0}{0.0} & \accstd{94.4}{1.4} & \accstd{95.0}{1.1} & \accstd{94.6}{1.7} & \accstd{95.2}{1.0} & \accstd{95.5}{0.8} \\
Qwen2.5 & 32B Instruct  & \accstd{95.0}{0.0} & \accstd{96.0}{1.4} & \accstd{96.0}{1.4} & \accstd{96.0}{1.8} & \accstd{95.7}{0.8} & \accstd{96.2}{1.3} \\
Qwen2.5 & 72B Instruct  & \accstd{93.2}{0.0} & \accstd{96.0}{0.5} & \accstd{96.0}{0.4} & \accstd{96.1}{0.8} & \accstd{96.0}{0.7} & \accstd{95.8}{0.3} \\
Gemma-4 & 2B IT         & \accstd{73.5}{0.0} & \accstd{88.2}{1.6} & \accstd{87.6}{2.1} & \accstd{89.3}{2.0} & \accstd{88.9}{2.5} & \accstd{89.5}{2.3} \\
Gemma-4 & 4B IT         & \accstd{82.3}{0.0} & \accstd{95.5}{4.2} & \accstd{96.3}{1.2} & \accstd{94.8}{2.0} & \accstd{96.7}{0.4} & \accstd{96.3}{1.7} \\
Gemma-4 & 31B IT        & \accstd{96.0}{0.0} & \accstd{97.3}{0.1} & \accstd{96.5}{0.4} & \accstd{96.8}{0.2} & \accstd{96.7}{0.4} & \accstd{97.0}{0.3} \\
Qwen3.5 & 2B            & \accstd{63.7}{0.0} & \accstd{78.6}{1.8} & \accstd{79.9}{1.6} & \accstd{80.9}{1.4} & \accstd{80.6}{1.8} & \accstd{79.8}{2.8} \\
Qwen3.5 & 9B            & \accstd{93.2}{0.0} & \accstd{97.9}{0.5} & \accstd{97.6}{0.3} & \accstd{97.5}{0.5} & \accstd{97.0}{0.3} & \accstd{97.1}{0.6} \\
Qwen3.5 & 27B           & \accstd{96.5}{0.0} & \accstd{98.0}{0.5} & \accstd{97.6}{0.3} & \accstd{97.7}{0.2} & \accstd{97.6}{0.3} & \accstd{97.6}{0.4} \\
\bottomrule
\end{tabular}}
\end{table*}

\begin{table*}[ht]
\centering
\scriptsize
\setlength{\tabcolsep}{3.2pt}
\renewcommand{\arraystretch}{1.08}
\caption{\textbf{Complete ProverQA results on the Medium split.}
Accuracy is reported as mean $\pm$ standard deviation over 5 random runs.}
\label{app:tab:proverqa_medium}
\resizebox{\textwidth}{!}{%
\begin{tabular}{llcccccc}
\toprule
\textbf{Family} & \textbf{Model} & \textbf{Zero-shot} & \textbf{0\%} & \textbf{25\%} & \textbf{50\%} & \textbf{75\%} & \textbf{100\%} \\
\midrule
Llama-3.1 & 8B Instruct  & \accstd{57.3}{0.0} & \accstd{72.9}{6.3} & \accstd{74.2}{3.9} & \accstd{71.0}{2.5} & \accstd{70.6}{2.9} & \accstd{68.1}{6.7} \\
Llama-3.1 & 70B Instruct & \accstd{74.3}{0.0} & \accstd{87.2}{1.5} & \accstd{88.8}{0.9} & \accstd{88.0}{1.3} & \accstd{87.0}{1.6} & \accstd{86.0}{1.5} \\
Qwen2.5 & 7B Instruct   & \accstd{69.3}{0.0} & \accstd{72.3}{0.8} & \accstd{73.9}{1.3} & \accstd{73.1}{0.3} & \accstd{70.7}{1.8} & \accstd{72.1}{1.1} \\
Qwen2.5 & 14B Instruct  & \accstd{83.0}{0.0} & \accstd{88.3}{0.8} & \accstd{87.0}{1.1} & \accstd{86.9}{0.6} & \accstd{86.7}{0.9} & \accstd{85.5}{0.8} \\
Qwen2.5 & 32B Instruct  & \accstd{79.7}{0.0} & \accstd{90.6}{0.7} & \accstd{89.5}{0.7} & \accstd{89.0}{1.0} & \accstd{88.7}{0.8} & \accstd{88.8}{0.9} \\
Qwen2.5 & 72B Instruct  & \accstd{82.0}{0.0} & \accstd{86.8}{0.9} & \accstd{86.0}{0.7} & \accstd{86.0}{1.0} & \accstd{85.8}{1.4} & \accstd{85.4}{1.3} \\
Gemma-4 & 2B IT         & \accstd{67.5}{0.0} & \accstd{76.1}{1.6} & \accstd{73.3}{2.9} & \accstd{73.7}{2.3} & \accstd{73.1}{3.3} & \accstd{70.2}{3.7} \\
Gemma-4 & 4B IT         & \accstd{65.5}{0.0} & \accstd{85.2}{0.5} & \accstd{84.1}{1.3} & \accstd{83.0}{0.6} & \accstd{82.0}{1.2} & \accstd{80.9}{1.0} \\
Gemma-4 & 31B IT        & \accstd{90.2}{0.0} & \accstd{91.5}{0.7} & \accstd{91.0}{0.6} & \accstd{90.5}{0.8} & \accstd{91.0}{0.8} & \accstd{90.8}{0.9} \\
Qwen3.5 & 2B            & \accstd{59.5}{0.0} & \accstd{69.3}{0.9} & \accstd{71.2}{3.1} & \accstd{70.7}{2.6} & \accstd{71.4}{2.0} & \accstd{70.2}{2.0} \\
Qwen3.5 & 9B            & \accstd{83.8}{0.0} & \accstd{86.9}{1.5} & \accstd{85.2}{1.0} & \accstd{85.4}{0.9} & \accstd{85.2}{2.0} & \accstd{85.2}{1.1} \\
Qwen3.5 & 27B           & \accstd{89.0}{0.0} & \accstd{92.1}{0.2} & \accstd{91.8}{0.3} & \accstd{91.9}{0.2} & \accstd{91.7}{0.6} & \accstd{91.7}{0.8} \\
\bottomrule
\end{tabular}}
\end{table*}

\begin{table*}[ht]
\centering
\scriptsize
\setlength{\tabcolsep}{3.2pt}
\renewcommand{\arraystretch}{1.08}
\caption{\textbf{Complete ProverQA results on the Hard split.}
Accuracy is reported as mean $\pm$ standard deviation over 5 random runs.}
\label{app:tab:proverqa_hard}
\resizebox{\textwidth}{!}{%
\begin{tabular}{llcccccc}
\toprule
\textbf{Family} & \textbf{Model} & \textbf{Zero-shot} & \textbf{0\%} & \textbf{25\%} & \textbf{50\%} & \textbf{75\%} & \textbf{100\%} \\
\midrule
Llama-3.1 & 8B Instruct  & \accstd{53.0}{0.0} & \accstd{57.8}{1.3} & \accstd{58.3}{2.4} & \accstd{55.5}{1.0} & \accstd{54.1}{2.4} & \accstd{51.9}{3.5} \\
Llama-3.1 & 70B Instruct & \accstd{58.3}{0.0} & \accstd{75.0}{2.3} & \accstd{73.7}{1.9} & \accstd{72.5}{1.9} & \accstd{70.6}{1.7} & \accstd{69.7}{2.5} \\
Qwen2.5 & 7B Instruct   & \accstd{55.0}{0.0} & \accstd{63.4}{1.9} & \accstd{62.6}{1.5} & \accstd{60.6}{1.1} & \accstd{60.9}{2.3} & \accstd{59.0}{2.4} \\
Qwen2.5 & 14B Instruct  & \accstd{56.5}{0.0} & \accstd{74.5}{1.7} & \accstd{71.1}{1.2} & \accstd{70.1}{1.7} & \accstd{68.6}{1.8} & \accstd{67.6}{1.4} \\
Qwen2.5 & 32B Instruct  & \accstd{62.5}{0.0} & \accstd{78.2}{1.0} & \accstd{75.1}{0.9} & \accstd{72.9}{1.4} & \accstd{73.4}{1.7} & \accstd{73.2}{0.8} \\
Qwen2.5 & 72B Instruct  & \accstd{59.3}{0.0} & \accstd{75.2}{0.8} & \accstd{74.3}{0.9} & \accstd{74.2}{0.9} & \accstd{72.0}{2.1} & \accstd{71.5}{1.8} \\
Gemma-4 & 2B IT         & \accstd{49.0}{0.0} & \accstd{52.6}{2.0} & \accstd{52.9}{1.2} & \accstd{51.8}{2.3} & \accstd{54.2}{3.0} & \accstd{54.1}{0.5} \\
Gemma-4 & 4B IT         & \accstd{61.5}{0.0} & \accstd{70.2}{0.9} & \accstd{69.7}{2.3} & \accstd{70.8}{1.5} & \accstd{69.1}{1.0} & \accstd{70.0}{0.7} \\
Gemma-4 & 31B IT        & \accstd{87.3}{0.0} & \accstd{85.9}{0.7} & \accstd{85.6}{0.9} & \accstd{85.6}{1.0} & \accstd{85.4}{0.3} & \accstd{84.6}{0.7} \\
Qwen3.5 & 2B            & \accstd{45.5}{0.0} & \accstd{44.7}{4.5} & \accstd{47.9}{3.4} & \accstd{48.6}{3.4} & \accstd{46.6}{2.6} & \accstd{45.8}{3.6} \\
Qwen3.5 & 9B            & \accstd{73.3}{0.0} & \accstd{71.5}{1.7} & \accstd{71.0}{2.3} & \accstd{72.3}{2.0} & \accstd{70.6}{2.1} & \accstd{69.7}{1.8} \\
Qwen3.5 & 27B           & \accstd{82.0}{0.0} & \accstd{87.1}{0.7} & \accstd{87.1}{0.9} & \accstd{86.3}{1.1} & \accstd{87.6}{0.9} & \accstd{86.6}{0.3} \\
\bottomrule
\end{tabular}}
\end{table*}
\clearpage

\subsection{All Results in Math Word Problems}
\label{app:all_math}

Table~\ref{app:tab:problemathic_appendix} reports the complete PROBLEMATHIC results on the Simple and Complex splits. 
All values are reported as Exact Match accuracy with mean $\pm$ standard deviation over 10 random runs, in percentage points. 

\begin{table*}[ht]
\centering
\scriptsize
\setlength{\tabcolsep}{4.5pt}
\caption{\textbf{Complete PROBLEMATHIC math word problem results.} 
Exact Match accuracy is reported as mean $\pm$ standard deviation over 10 random runs, in percentage points. 
The Simple and Complex splits are evaluated under different ratios of answer-preserving exemplar perturbations, where irrelevant numerical information is added to selected in-context exemplars while preserving the original solution and final answer.}
\label{app:tab:problemathic_appendix}
\resizebox{\linewidth}{!}{
\begin{tabular}{llccccc}
\toprule
Split & Model & 0\% & 25\% & 50\% & 75\% & 100\% \\
\midrule
Simple & LLAMA-2-7B  
& $79.0{\scriptsize\pm0.8}$ & $78.3{\scriptsize\pm1.3}$ & $77.9{\scriptsize\pm1.9}$ & $76.0{\scriptsize\pm2.7}$ & $69.6{\scriptsize\pm2.8}$ \\
Simple & LLAMA-2-13B 
& $77.4{\scriptsize\pm3.9}$ & $76.7{\scriptsize\pm3.6}$ & $77.1{\scriptsize\pm3.8}$ & $79.3{\scriptsize\pm1.7}$ & $78.9{\scriptsize\pm1.7}$ \\
Simple & LLAMA-2-70B 
& $73.5{\scriptsize\pm4.8}$ & $73.8{\scriptsize\pm4.5}$ & $73.5{\scriptsize\pm2.6}$ & $72.2{\scriptsize\pm4.0}$ & $67.1{\scriptsize\pm3.5}$ \\
\midrule
Complex & LLAMA-2-7B  
& $50.5{\scriptsize\pm2.8}$ & $46.8{\scriptsize\pm4.3}$ & $45.1{\scriptsize\pm4.5}$ & $37.9{\scriptsize\pm8.6}$ & $28.2{\scriptsize\pm5.7}$ \\
Complex & LLAMA-2-13B 
& $55.9{\scriptsize\pm3.0}$ & $56.6{\scriptsize\pm5.1}$ & $56.0{\scriptsize\pm5.4}$ & $56.2{\scriptsize\pm6.1}$ & $56.1{\scriptsize\pm5.5}$ \\
Complex & LLAMA-2-70B 
& $53.4{\scriptsize\pm4.9}$ & $53.5{\scriptsize\pm5.0}$ & $53.5{\scriptsize\pm5.2}$ & $56.9{\scriptsize\pm5.1}$ & $55.9{\scriptsize\pm6.4}$ \\
\bottomrule
\end{tabular}
}
\end{table*}

\subsection{All Results in Positional Effects}
\label{app:all_position}

We provide the full SST-2 positional-effect results in Tables~\ref{tab:sst2_appendix_llama} and~\ref{tab:sst2_appendix_qwen}. 
For each model, we compare five perturbation placement policies under the same replacement budget: random, middle, head, tail, and custom. The value outside parentheses denotes the accuracy on the normal SST-2 split, while the value inside parentheses denotes the signed accuracy change relative to the random-placement baseline under the same replacement budget. For each model and replacement budget, the lowest accuracy across placement policies is highlighted in \textbf{bold}, and the second lowest accuracy is \underline{underlined}. These detailed results supplement the main-text analysis by showing that the effect of perturbation placement is model-dependent: smaller models are more sensitive to where perturbed exemplars appear, whereas larger models remain comparatively stable across placement policies.

\begin{table*}[ht]
\centering
\caption{All SST-2 positional-effect results (\%) with 32 exemplars for Llama-family models.}
\label{tab:sst2_appendix_llama}
\scriptsize
\setlength{\tabcolsep}{3.5pt}
\renewcommand{\arraystretch}{1.08}
\begin{adjustbox}{width=\textwidth}
\begin{tabular}{cc|ccccc}
\toprule
\multirow{2}{*}{Replaced} & \multirow{2}{*}{Method}
& \multicolumn{3}{c}{Llama-2}
& \multicolumn{2}{c}{Llama-3.1} \\
\cmidrule(lr){3-5} \cmidrule(lr){6-7}
& & 7B-chat & 13B-chat & 70B-chat & 8B-Instruct & 70B-Instruct \\
\midrule

\multirow{5}{*}{8}
& random & 98.5 {\scriptsize (+0.0)} & \underline{99.9} {\scriptsize (+0.0)} & \textbf{99.4} {\scriptsize (+0.0)} & \underline{98.1} {\scriptsize (+0.0)} & \textbf{100.0} {\scriptsize (+0.0)} \\
& middle & \underline{98.2} {\scriptsize (-0.3)} & \textbf{99.8} {\scriptsize (-0.1)} & 99.6 {\scriptsize (+0.3)} & 99.7 {\scriptsize (+1.6)} & \textbf{100.0} {\scriptsize (+0.0)} \\
& head   & 98.8 {\scriptsize (+0.4)} & 100.0 {\scriptsize (+0.1)} & \underline{99.5} {\scriptsize (+0.1)} & \textbf{97.9} {\scriptsize (-0.2)} & \textbf{100.0} {\scriptsize (+0.0)} \\
& tail   & \textbf{97.4} {\scriptsize (-1.1)} & \textbf{99.8} {\scriptsize (-0.1)} & 99.6 {\scriptsize (+0.2)} & 99.7 {\scriptsize (+1.6)} & \textbf{100.0} {\scriptsize (+0.0)} \\
& custom & 98.5 {\scriptsize (+0.0)} & \textbf{99.8} {\scriptsize (-0.1)} & 99.6 {\scriptsize (+0.2)} & 99.5 {\scriptsize (+1.4)} & \textbf{100.0} {\scriptsize (+0.0)} \\
\midrule

\multirow{5}{*}{16}
& random & 94.7 {\scriptsize (+0.0)} & 99.0 {\scriptsize (+0.0)} & 99.7 {\scriptsize (+0.0)} & \underline{96.9} {\scriptsize (+0.0)} & \underline{100.0} {\scriptsize (+0.0)} \\
& middle & 94.3 {\scriptsize (-0.5)} & \underline{98.7} {\scriptsize (-0.3)} & \underline{99.6} {\scriptsize (-0.1)} & 98.7 {\scriptsize (+1.8)} & \textbf{99.9} {\scriptsize (-0.1)} \\
& head   & 95.2 {\scriptsize (+0.4)} & \textbf{98.1} {\scriptsize (-0.9)} & \textbf{99.4} {\scriptsize (-0.4)} & \textbf{91.5} {\scriptsize (-5.5)} & \textbf{99.9} {\scriptsize (-0.1)} \\
& tail   & \textbf{92.3} {\scriptsize (-2.5)} & 99.0 {\scriptsize (+0.0)} & \underline{99.6} {\scriptsize (-0.1)} & 98.4 {\scriptsize (+1.5)} & \underline{100.0} {\scriptsize (+0.0)} \\
& custom & \underline{92.5} {\scriptsize (-2.2)} & 98.9 {\scriptsize (-0.1)} & 99.7 {\scriptsize (+0.0)} & 97.4 {\scriptsize (+0.5)} & \underline{100.0} {\scriptsize (+0.0)} \\
\midrule

\multirow{5}{*}{24}
& random & 88.2 {\scriptsize (+0.0)} & 99.7 {\scriptsize (+0.0)} & 99.7 {\scriptsize (+0.0)} & 94.9 {\scriptsize (+0.0)} & \textbf{100.0} {\scriptsize (+0.0)} \\
& middle & \textbf{74.8} {\scriptsize (-13.4)} & \underline{99.5} {\scriptsize (-0.3)} & 99.2 {\scriptsize (-0.5)} & 95.9 {\scriptsize (+1.0)} & \textbf{100.0} {\scriptsize (+0.0)} \\
& head   & 81.1 {\scriptsize (-7.1)} & \textbf{99.0} {\scriptsize (-0.7)} & 99.2 {\scriptsize (-0.5)} & \textbf{83.4} {\scriptsize (-11.6)} & \textbf{100.0} {\scriptsize (+0.0)} \\
& tail   & \underline{77.5} {\scriptsize (-10.6)} & 99.6 {\scriptsize (-0.1)} & \underline{98.5} {\scriptsize (-1.3)} & 97.1 {\scriptsize (+2.2)} & \textbf{100.0} {\scriptsize (+0.0)} \\
& custom & 78.0 {\scriptsize (-10.2)} & \underline{99.5} {\scriptsize (-0.3)} & \textbf{96.3} {\scriptsize (-3.5)} & \underline{94.4} {\scriptsize (-0.6)} & \textbf{100.0} {\scriptsize (+0.0)} \\
\midrule

\multirow{5}{*}{28}
& random & 72.9 {\scriptsize (+0.0)} & \textbf{98.6} {\scriptsize (+0.0)} & 99.5 {\scriptsize (+0.0)} & \underline{88.5} {\scriptsize (+0.0)} & \textbf{99.9} {\scriptsize (+0.0)} \\
& middle & 61.0 {\scriptsize (-11.9)} & 99.2 {\scriptsize (+0.6)} & \underline{97.4} {\scriptsize (-2.1)} & 91.8 {\scriptsize (+3.3)} & \underline{100.0} {\scriptsize (+0.1)} \\
& head   & 70.1 {\scriptsize (-2.8)} & \textbf{98.6} {\scriptsize (+0.0)} & 98.2 {\scriptsize (-1.3)} & \textbf{85.2} {\scriptsize (-3.4)} & \textbf{99.9} {\scriptsize (+0.0)} \\
& tail   & \textbf{58.1} {\scriptsize (-14.8)} & \underline{98.8} {\scriptsize (+0.3)} & 98.3 {\scriptsize (-1.2)} & 92.8 {\scriptsize (+4.3)} & \underline{100.0} {\scriptsize (+0.1)} \\
& custom & \underline{59.6} {\scriptsize (-13.3)} & \textbf{98.6} {\scriptsize (+0.0)} & \textbf{97.2} {\scriptsize (-2.3)} & 90.6 {\scriptsize (+2.1)} & \underline{100.0} {\scriptsize (+0.1)} \\
\bottomrule
\end{tabular}
\end{adjustbox}
\end{table*}

\begin{table*}[ht]
\centering
\caption{All SST-2 positional-effect results (\%) with 32 exemplars for Qwen2.5-family models.}
\label{tab:sst2_appendix_qwen}
\scriptsize
\setlength{\tabcolsep}{3.5pt}
\renewcommand{\arraystretch}{1.08}
\begin{adjustbox}{width=\textwidth}
\begin{tabular}{cc|ccccc}
\toprule
\multirow{2}{*}{Replaced} & \multirow{2}{*}{Method}
& \multicolumn{5}{c}{Qwen2.5} \\
\cmidrule(lr){3-7}
& & 3B-Instruct & 7B-Instruct & 14B-Instruct & 32B-Instruct & 72B-Instruct \\
\midrule

\multirow{5}{*}{8}
& random & 97.9 {\scriptsize (+0.0)} & 99.6 {\scriptsize (+0.0)} & \underline{99.9} {\scriptsize (+0.0)} & 100.0 {\scriptsize (+0.0)} & \underline{100.0} {\scriptsize (+0.0)} \\
& middle & 97.8 {\scriptsize (-0.1)} & 99.3 {\scriptsize (-0.3)} & 100.0 {\scriptsize (+0.1)} & 100.0 {\scriptsize (+0.0)} & \textbf{99.9} {\scriptsize (-0.1)} \\
& head   & \textbf{95.7} {\scriptsize (-2.2)} & \textbf{98.8} {\scriptsize (-0.7)} & 100.0 {\scriptsize (+0.1)} & 100.0 {\scriptsize (+0.0)} & \textbf{99.9} {\scriptsize (-0.1)} \\
& tail   & \underline{97.1} {\scriptsize (-0.8)} & 99.6 {\scriptsize (+0.0)} & \underline{99.9} {\scriptsize (+0.0)} & \textbf{99.7} {\scriptsize (-0.3)} & \textbf{99.9} {\scriptsize (-0.1)} \\
& custom & 97.5 {\scriptsize (-0.5)} & \underline{99.0} {\scriptsize (-0.6)} & \textbf{99.6} {\scriptsize (-0.3)} & \underline{99.9} {\scriptsize (-0.1)} & \textbf{99.9} {\scriptsize (-0.1)} \\
\midrule

\multirow{5}{*}{16}
& random & 98.2 {\scriptsize (+0.0)} & \textbf{98.6} {\scriptsize (+0.0)} & 99.8 {\scriptsize (+0.0)} & 99.8 {\scriptsize (+0.0)} & \underline{99.8} {\scriptsize (+0.0)} \\
& middle & \underline{95.7} {\scriptsize (-2.4)} & 99.2 {\scriptsize (+0.5)} & \underline{98.8} {\scriptsize (-1.0)} & 99.7 {\scriptsize (-0.1)} & 99.9 {\scriptsize (+0.1)} \\
& head   & \textbf{89.0} {\scriptsize (-9.2)} & \underline{98.9} {\scriptsize (+0.3)} & 99.8 {\scriptsize (+0.0)} & \underline{99.6} {\scriptsize (-0.2)} & \textbf{99.2} {\scriptsize (-0.6)} \\
& tail   & 96.2 {\scriptsize (-2.0)} & 99.3 {\scriptsize (+0.6)} & \textbf{98.6} {\scriptsize (-1.2)} & \textbf{98.8} {\scriptsize (-1.0)} & 100.0 {\scriptsize (+0.2)} \\
& custom & 97.3 {\scriptsize (-0.9)} & \underline{98.9} {\scriptsize (+0.3)} & 99.4 {\scriptsize (-0.5)} & \underline{99.6} {\scriptsize (-0.2)} & 99.9 {\scriptsize (+0.1)} \\
\midrule

\multirow{5}{*}{24}
& random & 95.0 {\scriptsize (+0.0)} & \underline{98.4} {\scriptsize (+0.0)} & 99.9 {\scriptsize (+0.0)} & \textbf{99.8} {\scriptsize (+0.0)} & 100.0 {\scriptsize (+0.0)} \\
& middle & \underline{93.5} {\scriptsize (-1.5)} & 99.1 {\scriptsize (+0.7)} & \underline{99.7} {\scriptsize (-0.2)} & \underline{100.0} {\scriptsize (+0.2)} & \underline{99.8} {\scriptsize (-0.2)} \\
& head   & \textbf{89.8} {\scriptsize (-5.2)} & \textbf{98.0} {\scriptsize (-0.4)} & \textbf{99.6} {\scriptsize (-0.4)} & \textbf{99.8} {\scriptsize (+0.0)} & \textbf{99.7} {\scriptsize (-0.3)} \\
& tail   & 95.4 {\scriptsize (+0.4)} & 99.2 {\scriptsize (+0.8)} & 99.8 {\scriptsize (-0.1)} & \underline{100.0} {\scriptsize (+0.2)} & \underline{99.8} {\scriptsize (-0.2)} \\
& custom & 95.6 {\scriptsize (+0.6)} & 99.2 {\scriptsize (+0.8)} & \underline{99.7} {\scriptsize (-0.2)} & \underline{100.0} {\scriptsize (+0.2)} & 100.0 {\scriptsize (+0.0)} \\
\midrule

\multirow{5}{*}{28}
& random & \textbf{93.1} {\scriptsize (+0.0)} & \textbf{96.1} {\scriptsize (+0.0)} & \underline{99.4} {\scriptsize (+0.0)} & \textbf{99.3} {\scriptsize (+0.0)} & 100.0 {\scriptsize (+0.0)} \\
& middle & 98.2 {\scriptsize (+5.1)} & 98.8 {\scriptsize (+2.7)} & 99.5 {\scriptsize (+0.1)} & 99.8 {\scriptsize (+0.5)} & \underline{99.8} {\scriptsize (-0.2)} \\
& head   & 93.6 {\scriptsize (+0.6)} & \underline{97.3} {\scriptsize (+1.2)} & \textbf{99.2} {\scriptsize (-0.2)} & \underline{99.6} {\scriptsize (+0.3)} & \textbf{99.7} {\scriptsize (-0.3)} \\
& tail   & 98.9 {\scriptsize (+5.8)} & 99.4 {\scriptsize (+3.3)} & 99.5 {\scriptsize (+0.1)} & 100.0 {\scriptsize (+0.7)} & 99.9 {\scriptsize (-0.1)} \\
& custom & \underline{93.5} {\scriptsize (+0.4)} & 98.8 {\scriptsize (+2.7)} & 99.5 {\scriptsize (+0.1)} & 100.0 {\scriptsize (+0.7)} & 100.0 {\scriptsize (+0.0)} \\
\bottomrule
\end{tabular}
\end{adjustbox}
\end{table*}